%% file: Neurips/main.tex
\documentclass{article}

\PassOptionsToPackage{numbers,sort&compress}{natbib}

\usepackage[preprint]{Neurips/neurips_2026}
\usepackage{amsmath,amssymb,amsfonts,amsthm}  
\usepackage{graphicx}
\usepackage{comment}

\usepackage[utf8]{inputenc} 
\usepackage[T1]{fontenc}    
\usepackage{hyperref}       
\usepackage{url}            
\usepackage{booktabs}       
\usepackage{amsfonts}       
\usepackage{nicefrac}       
\usepackage{microtype}      
\usepackage{xcolor}         
\usepackage{algorithm}
\usepackage{algpseudocode}
\usepackage{mdframed}
\usepackage{enumitem}

\newcommand{\gi}[1]{\textcolor{blue}{[g:#1]}}
\newcommand{\ma}[1]{\textcolor{red}{[m:#1]}}
\newtheorem{assumption}{Assumption}

\newtheorem{theorem}{Theorem}

\newtheorem{lemma}{Lemma}
\newtheorem{corollary}{Corollary}
\newtheorem{remark}{Remark}

\usepackage{amsmath,amssymb,amsthm}

\usepackage{booktabs,multirow,colortbl,xcolor} 
\usepackage{pifont} 
\usepackage{tikz}
\usetikzlibrary{arrows.meta,positioning,calc,decorations.pathreplacing}
\newcommand{\cmark}{\ding{51}}  
\newcommand{\xmark}{\ding{55}}  

\title{Dynamic Optimistic Constrained OCO with Memory via Delay Equivalence}

\author{%
  Mohammed Abdullah \\
  L2S, CentraleSup\'elec \\
  Universit\'e Paris-Saclay, CNRS \\
  France \\
  \texttt{\small mohammed.abdullah@centralesupelec.fr}
  \And
  George Iosifidis \\
  Delft University of Technology \\
  The Netherlands \\
  \texttt{\small G.Iosifidis@tudelft.nl}
  \AND
  Salah Eddine Elayoubi \\
  L2S, CentraleSup\'elec \\
  Universit\'e Paris-Saclay, CNRS \\
  France \\
  \texttt{\small salaheddine.elayoubi@centralesupelec.fr}
  \And
  Tijani Chahed \\
  T\'el\'ecom SudParis \\
  Institut Polytechnique de Paris \\
  France \\
  \texttt{\small tijani.chahed@telecom-sudparis.eu}
}

\begin{document}

\maketitle

\begin{abstract}

Online Convex Optimization with memory and time-varying constraints (COCO-M) is a recently-introduced framework that captures the dynamics of stateful online learning and non-stochastic control with budget constraints. Despite its expressive power, COCO-M remains largely understudied. In this work, we tackle this problem in its full scale, providing an optimistic algorithm that incorporates untrusted predictions to tame universal dynamic regret and cumulative constraint violation. Our analysis leverages a potent reduction from memory to delay and treats state-dependent gradients as delayed feedback. We establish the first bounds that adapt to both environment non-stationarity (path length $P_T$) and prediction accuracy, while lifting previous restrictive assumptions such as functional separability. En route to these results, we derive findings of independent interest for the memory-to-delay reduction, and for delayed OCO with time-varying constraints. The proposed framework ensures sublinear worst-case guarantees that improve with prediction accuracy, reaching \(\mathcal O(1)\) regret under perfect predictions, with cumulative constraint violation (CCV) $\mathcal O(\sqrt{T})$ for unknown $P_T$ and $\mathcal O(\log T)$ for known or online observable $P_T$ , while recovering the optimal COCO rates when memory disappears.

\end{abstract}

\input{Neurips/Intro}

\input{Neurips/preliminaries}

\input{Neurips/OCO}

\input{Neurips/COCO}

\input{Neurips/memory}

\input{Neurips/Conclusion}

\newpage
\section*{Acknowledgments}
This work was supported in part by the French government through the France 2030 program under the Celtic RAI-6Green project, and in part by the European Commission under Grant No. 101192462 (FLECON-6G).



\bibliographystyle{unsrtnat}
\bibliography{citation} 
\newpage

\input{Neurips/appendix}

\end{document}

%% file: Neurips/Intro.tex
\section{Introduction}

Online Convex Optimization (OCO) is a powerful framework for sequential decision making: at each round $t\!=\!1\ldots, T$, a learner selects $x_t \in \mathcal X$, an adversary reveals a convex loss function $f_t:\mathcal X \to \mathbb{R}$, and the learner incurs cost $f_t(x_t)$. Performance is measured via regret, $\mathcal R_T$, which compares the learner’s cumulative loss against a benchmark sequence $\{u_t\}_t$. Depending on the choice of benchmark, one obtains static regret against a fixed decision $u_t=x^\star, \forall t$; worst-case dynamic regret against per-round minimizers, $u_t=x_t^\star=\arg\min_{x\in\mathcal X} f_t(x)$, $\forall t$; or the more general \emph{universal} dynamic regret, which allows comparison to any sequence $\{u_t\}_t$. Over the years, OCO has been extended to a wide range of settings with many practical applications \cite{hazan2016introduction,orabona2019modern}.


A key extension is OCO with memory (OCO-M), where the loss at time $t$ depends on a window of past decisions $x_{t-m}, \ldots, x_t$. This formulation arises naturally in applications such as recommender systems, portfolio optimization, and continual learning \cite{anava2015online}. Importantly, OCO-M underlies the framework of non-stochastic control (NSC), 
which optimizes stateful systems under adversarial disturbances \cite{hazan2026introductiononlinecontrol}. In many such problems, the learner must also satisfy time-average constraints of adversarially-varying functions $g_t(x_t)\leq 0$. In this \emph{constrained} OCO (COCO) setting, the goal is to control both regret and cumulative constraint violation (CCV), $Q_T \doteq \sum_{t} \max\{g_t(x_t),0\}$. In practice, these constraints often exhibit memory as well, depending jointly on past decisions. Following \cite{Abdullah_Iosifidis_Elayoubi_Chahed_2026}, we refer to this setting as COCO-M.



Despite its many applications in stateful online decision problems and NSC with time-average constraints, COCO-M remains largely unexplored. This gap is not due to a lack of attention, but reflects inherent technical barriers. Indeed, even simplified versions of this problem challenge existing OCO methods: memory induces delay between decisions and their fully-observed feedback, time-varying constraints interact with long-term feasibility requirements, and dynamic benchmarks introduce additional non-stationarity. As a result, current techniques are insufficient to address COCO-M in its full generality, motivating the development of new methodological approaches.


\textbf{Prior Work.} We review briefly here the main lines of pertinent work and refer the reader to Appendix \ref{sec:appendix-related} for a comprehensive discussion. While COCO attracts growing attention, see \cite{sinha2025beyond} and references therein, only few works consider dynamic regret. Under this benchmark, the learner must additionally adapt to the variability of comparator, measured by its path length $P_T \!=\! \sum_t \|u_t\! -\! u_{t-1}\|$. \cite{yi2021regret} established worst-case dynamic regret $\mathcal{O}(\sqrt{(1+P_T)T})$ and $Q_T\!=\!\mathcal{O}(\sqrt{T})$ for time-invariant constraints. For time-varying adversarial constraints, \cite{guo2022online} obtained $\mathcal{O}(\sqrt{(1\!+\!P_T)T})$ universal dynamic regret and $Q_T\!=\!\mathcal{O}(T^{3/4})$. Subsequent works \cite{lekeufack2024optimistic,liu2021simultaneously} refined these guarantees under different assumptions, and \cite{supantha2025universal} provided the first bounds for universal dynamic regret and CCV. However, none of these important works considers memory. The few studies of dynamic regret in memory-dependent problems are confined to the special case of switching costs (one-round memory) \cite{zhang2021revisiting}, \cite{ijcai2025p758}, or to NSC without time-average constraints \cite{zhao2023non}, \cite{baby2022optimal}, \cite{10384153}. Thus, dynamic regret for COCO-M, and more broadly for constrained NSC, remains essentially open.


From a methodological standpoint, this work brings together memory, delay, and
optimism. Existing static optimistic memory-to-delay reductions
\cite{mhaisen2024optimistic,Abdullah_Iosifidis_Elayoubi_Chahed_2026} require
separability across memory variables, which in COCO-M yields a restrictive
benchmark and prevents a direct application to constrained NSC; we drop this assumption by introducing a more general memory-to-delay reduction. Learning with delay is typically treated with multiple interleaved learners, e.g., \cite{joulani2013online}, while \cite{flaspohler2021online} unified delay and optimism by interpreting the delayed gradients as ''bad hints''. This line of work focuses on static regret, with few exceptions such as \cite{wan2024non} which does not consider constraints. Finally, optimistic learning, i.e., incorporating potentially inaccurate predictions in OCO, has been studied with dynamic regret in \cite{scroccaro2023adaptive} and \cite{mhaisen2025dynamic}, but without delay or memory effects. In summary, current methods do not provide the tools required to address even intermediate variants of the problem considered here.




\textbf{Contributions.} Motivated by this gap, we study COCO-M under universal dynamic regret while leveraging (possibly inaccurate) predictions of the losses and constraints. We propose a meta-learning framework that combines an adaptive hedge-based meta-learner with a collection of optimistic delayed Online Gradient Descent (OGD) base learners, each tuned to a different value of the (unknown) path length $P_T$. Both the meta-learner and the base learners are bespoke. Moreover, a key idea in our work is that learning with memory can be effectively mapped to learning with delay; we formalize this connection and employ it to study the targeted COCO-M problem. To the best of our knowledge, this is the first work to address this problem, and its contributions can be summarized as follows:
\begin{itemize}[leftmargin=0.5cm]
    
    \item Develops an optimistic delayed OCO algorithm and derives prediction-dependent universal dynamic regret guarantees (Theorem \ref{thm:delayed_optimistic_oco}). This is the algorithm engine of the framework.
    
    \item Extends this framework to constrained OCO with time-varying constraints, obtaining regret and cumulative constraint violation bounds for unknown and known $P_T$ (Theorems \ref{thm:unknown-path-quadratic-ccv} and \ref{thm:known-PT}).
    
    \item Establishes a reduction from memory-dependent OCO to delayed OCO via partial gradient aggregation, without imposing function separability (Lemma \ref{lem:memory_to_delay}). Building on this, it employs delayed OCO to deliver universal dynamic regret and CCV bounds for COCO-M (Corollary \ref{coro:memory_to_delay}).
\end{itemize}
In all cases, our optimistic COCO-M bounds subsume the best-known results of simpler problem versions, e.g., without memory (or delay) and/or without predictions; and, when dropping the optimism, we obtain fully-adaptive COCO-M bounds, covering an important gap in literature.  The details of proofs and intermediate results (some of independent interest) are included in Appendices.

%% file: Neurips/preliminaries.tex
\section{Background \& Preliminaries}\label{sec:prelimi}

\textbf{Notation.}
Let \(\mathcal{X}\!\subset\!\mathbb{R}^{d}\) be a nonempty, closed, bounded and convex set, with \(\ell_2\)-norm diameter \(D \doteq \sup_{x,y\in\mathcal{X}} \|x\!-\!y\|\). We write \([T]:=\{1,\ldots,T\}\) and, when needed, 
\( [T_m]:=\{m,\ldots,T\}\), where $m$ is the memory window, and use the shorthand notation:
\begin{equation}
x_{t-m}^t\;\doteq\;(x_{t-m},\dots,x_t)\in\mathcal{X}^{m+1}, \quad x_{a:b}=\sum_{i=a}^{b}x_i, \quad \text{and} \quad z^+\doteq \max\{z,0\}, \ z\in \mathbb R. \notag
\end{equation}
We denote with \(\tilde f_t\) the prediction of the function \(f_t\) and with \(\nabla_{x_s} f_t(\cdot)\) the partial gradient (a subgradient) of \(f_t\) with respect to \(x_s\). Also, $\Delta_N$ is the $N$-dimensional unit simplex.

\textbf{Metrics \& Benchmarks.}
At each round \(t\!\in\! [T_m]\), the learner decides \(x_t\!\in\!\mathcal X\) and incurs loss \(f_t(x_{t-m}^t)\) and constraint violation \(g_t(x_{t-m}^t)\), where
\(
f_t,g_t:\mathcal X^{m+1}\!\mapsto\!\mathbb R.
\)
We compare the learner against a comparator sequence
\(
u_{1}^T\doteq (u_1,u_2,\ldots,u_T)
\)
that belongs to the feasible benchmark set:
\[
\mathcal U_T
:=
\left\{
u_{1}^T\in \mathcal X^T
:
g_t\bigl(u_{t-m}^t\bigr)\le 0,\ \forall t\in [T_m]
\right\},
\]
where we use the convention $x_0\!=\!u_0\!=\!0$, to streamline presentation. Observe that $\mathcal U_T$ is confined to satisfy the constraints on a per-round basis. This is necessary so as to avoid the impossibility result of \cite{mannor2009online} and is in line with prior COCO works (see Appendix Sec. \ref{sec:appendix-related}). Still, $\mathcal U_T$ expands the static-regret COCO-M benchmark set in \cite{Abdullah_Iosifidis_Elayoubi_Chahed_2026}; an advantage of our new memory-to-delay reduction in Sec. \ref{sec:memory_delay}.

The learner is assessed using the metrics of dynamic regret and cumulative constraint violation\footnote{If there are multiple constraints \(g_{t,k}\), \(k=1,\dots,n\), one may define \(g_t(\cdot)\doteq \max_{k\in[n]} g_{t,k}(\cdot)\).}:
\begin{align}
\mathcal{R}_T(f)\doteq \sum_{t=m}^{T} f_t(x_{t-m}^{t})-\sum_{t=m}^{T} f_t(u_{t-m}^{t}), \qquad 
\mathrm{Q_T}\doteq \sum_{t=m}^{T} g_t^{+}(x_{t-m}^{t}).
\end{align}
Our goal is to ensure both \(\mathcal{R}_T(f)\) and \(Q_T\) are sublinear in \(T\) and the comparator's path length \(P_T\doteq \sum_{t=1}^{T-1}\|u_{t+1}-u_t\|\). Here we study \emph{universal} dynamic regret, which is measured against all feasible comparator sequences \cite{zhang2018dynamic}; this is a more general (and more challenging) metric \cite{supantha2025universal}.


\textbf{Meta-learning with Hedge}. Optimizing for dynamic regret requires knowledge of $P_T$ or adaptation at runtime. A key idea here is to use a meta-learner that hedges against a set of $N\!=\!\left\lceil \frac{1}{2}\log_2(1+DT)\right\rceil+1$ base learners, that are implemented with different learning rates in a way that ensures at least one of them is tuned within a constant factor of \(\sqrt{1+P_T}\) \cite{zhang2018adaptive}. Unfortunately, in our problem the hedging is compounded by the memory effects that introduce delay in discerning the best base learner.




\textbf{Memory-to-Delay}. In memory-less OCO, the loss gradient is available immediately; however, in OCO-M the impact of \(x_t\) is dispersed across the next \(m\) rounds; thus, memory induces feedback delay. To state this connection in a form that also applies to constrained problems, let \(L_t\!:\!\mathcal X^{m+1}\!\mapsto \mathbb R\) be a memory-dependent loss. In the unconstrained case, \(L_t=f_t\), while in the constrained case \(L_t\) is the penalized surrogate loss. We define the \emph{credit gradient} \(h_s\) as the sum of partial gradients of all functions influenced by \(x_s\), i.e.,
\(
h_s \!=\! \sum_{t\in I_s}\nabla_{x_s} L_t(x_{t-m}^t)
\)\footnote{We use \(\nabla\) throughout for simplicity; if the functions are not differentiable, \(\nabla\) should be interpreted as a subgradient \(\partial\)}, where $I_s\!=\!\{t\in [T_m],t-m\!\leq\! s\!\leq\! t\}$.
Clearly, \(h_s\) cannot be computed until round \(s+m\), which reveals a feedback delay of \(m\) rounds in OCO-M. Using this observation (formalized in Sec.~\ref{sec:memory_delay}), we shift our focus from the per-round loss \(L_t\) to the per-decision credit \(h_s\), i.e., from memory to a delayed feedback problem. A similar transformation, at the function level, was used by \cite{Abdullah_Iosifidis_Elayoubi_Chahed_2026} and \cite{mhaisen2024optimistic}; we discuss the differences in Sec.~\ref{sec:memory_delay}. 

\textbf{Optimism}. Another key idea in this work is optimism, i.e., incorporating in learning (possibly inaccurate) predictions of future gradient vectors and/or actions \cite{rakhlin2013online}. Optimistic learning turns non-optimistic regret bounds of the form $\mathcal O(T^k)$, for some $k\in[0,1]$, to $\mathcal O(E_T^k)$, where $E_T$ is the accumulated prediction error. These bounds hence range from $\mathcal O(1)$ for accurate predictions, to the non-optimistic bound $\mathcal O(T^k)$ in the worse case. This improvement happens, typically, through prediction errors-adaptive regularization, e.g., see \cite{mohri2016accelerating, 10.1016/j.tcs.2019.11.015}, and do not require prior information of the prediction accuracy. Finally, an important connection between optimism and feedback delay in OCO, which we also leverage here, was recently introduced in \cite{flaspohler2021online}.

We rely on the following standard OCO assumptions throughout the paper.
\begin{assumption}\label{ass:domain}
The decision set \(\mathcal{X}\) is nonempty, closed, bounded, and convex,
with \({\bf 0}\in\mathcal{X}\) and diameter \(D\).
\end{assumption}
\begin{assumption}\label{ass:convex}
For every \(t\in[T_m]\), the functions
\(f_t,g_t:\mathcal{X}^{m+1}\to\mathbb{R}\) are convex in their arguments
and uniformly bounded in absolute value by \(G\).
\end{assumption}
\begin{assumption}\label{ass:lipsh}
For every \(t\in[T_m]\), the gradients of \(f_t\) and \(g_t\), as well as
their predicted counterparts, are uniformly bounded in norm by \(G\).
\end{assumption}

Finally, we consider an oblivious adversary, in line with \cite{anava2015online,merhav2002sequential,gyorgy2014near} and others. Hence, the loss and constraint functions are determined at $t\!=\!0$, but are not revealed before each round $t$ is realized.
\begin{remark}
Unlike prior optimistic OCO/COCO-M works \cite{Abdullah_Iosifidis_Elayoubi_Chahed_2026,mhaisen2024optimistic}, Assumptions \ref{ass:domain}-\ref{ass:lipsh} do not impose separability of the memory-dependent functions. Further, we do not impose Slater-type conditions.
\end{remark}


\textbf{Roadmap.} Given the identified memory-to-delay equivalence, we start in Section~\ref{sec:OCO} by studying universal dynamic regret for the optimistic delayed OCO problem. In Section~\ref{sec:COCO}, we extend this analysis to the case of time-varying adversarial constraints (COCO with delay). Finally, Section~\ref{sec:memory_delay} formalizes the memory-to-delay connection and shows how these results can be transferred to the targeted optimistic COCO-M problem.

%% file: Neurips/OCO.tex
\section{Dynamic Optimistic OCO with Delay}\label{sec:OCO}


We start with the problem of optimistic OCO with feedback delay. We use a decision-time convention that makes the delay explicit and facilitates making later the connection with memory. The decision \(x_t\) is played at time \(t\), while its associated feedback vector \(h_t\) is fully revealed only after \(m\) rounds. Equivalently, at round \(t\), the learner receives the delayed feedback \(h_{t-m}\). Abusing slightly the notation (in order to facilitate presentation in Section~\ref{sec:memory_delay}), we redefine \(L_{t+m}\!:\!\mathcal X\mapsto\!\mathbb R\) to denote the convex loss at round \(t+m\) which is associated (due to delay) with the earlier decision \(x_t\). We also redefine \(h_t\in\nabla L_{t+m}(x_t)\).
This notation aims to emphasize the delayed feedback. By convexity:
\[
L_{t+m}(x_t)-L_{t+m}(u_t)
\le
\langle h_t,x_t-u_t\rangle, \quad \forall \ u_{1}^{T}\in\mathcal X^T
\]
Thus, we control the regret \(\mathcal R_T(\ell):=\sum_{t=1}^{T}\langle h_t,x_t-u_t\rangle\) of the delayed linear losses \(\ell_t(x):=\langle h_t,x\rangle\).



We study the challenging case where the comparator path length \(P_T\) is unknown, and employ a meta-learning construction with $N$ base learners, similarly in spirit to \cite{zhang2018adaptive}. At each round \(t\), every base learner \(i\in[N]\) proposes an action \(x_t^{(i)}\!\in\!\mathcal X\), and the meta-learner plays a \emph{fused} action:
\[
x_t=X_t w_t,
\qquad \text{where }
X_t:=[x_t^{(1)},\dots,x_t^{(N)}]
\qquad \text{and } \ \ \ \
w_t\in\Delta_N .
\]
For any fixed base learner \(i\in[N]\), the delayed linearized regret decomposes as:
\[
\mathcal{R}_T(\ell)
=
\underbrace{
\sum_{t=1}^{T}\langle h_t,x_t-x_t^{(i)}\rangle
}_{\textnormal{Term (I)}}
+
\underbrace{
\sum_{t=1}^{T}\langle h_t,x_t^{(i)}-u_t\rangle
}_{\textnormal{Term (II)}} .
\]
Since \(x_t=X_tw_t\) and \(x_t^{(i)}=X_te_i\), where \(e_i\) is the \(i\)-th vertex of
\(\Delta_N\), Term~(I) becomes:
\[
\textnormal{Term (I)}
=
\sum_{t=1}^{T}\langle M_t,w_t-e_i\rangle,
\qquad \text{with} \qquad
M_t:=X_t^\top h_t .
\]
This way, we obtain a meta-level problem with losses
\(M_t\) and a base-level problem with losses \(h_t\); and we treat both problems with delayed learning and optimism. 

Unfortunately, standard dynamic-regret meta-learning schemes such as
\cite{zhang2018adaptive} do not handle delayed expert losses, while
delayed/adaptive Hedge methods such as
\cite{korotin2020adaptive,orabona2019modern} do not provide the optimistic prediction-dependent form needed here. Therefore, we build a bespoke meta-learning construction, leveraging and extending the delay-as-optimism framework of \cite{flaspohler2021online}: the missing delayed gradients are handled through predictions (or, \emph{hints}) rather than worst-case gradient bounds; and we use a delayed-upper-bound (DUB) tuning to set the regularization from revealed quantities, thus avoiding over-regularization and preserving the benefit of predictions.

Let us clarify first the prediction structure which is essentially common for both problems. At each round \(t\), after observing \(h_{t-m}\), the block
\(h_{t-m+1},\ldots,h_t\) remains unavailable to the learner. To overcome this issue, the algorithm uses the prediction  block
\(\widetilde h_{t-m+1}^{t+1}\), which includes both this missing information and the next hint \(\widetilde h_{t+1}\). That is, at each round $t$, the learner needs to predict the gradient of $L_{t+1}, \ldots, L_{t+m+1}$, and also the $x_{t+1}$, as it appears in the function of round $(t+m+1)$. We emphasize that this is not the only possible way to construct the hint. As in
prior optimistic works, predicted gradients may be evaluated instead at current decisions (which are known); see
\cite{flaspohler2021online,scroccaro2023adaptive,rakhlin2013online}. More generally, the algorithm does not require these quantities to be predicted separately: it only needs a bounded vector \(\widetilde h_s\) that serves as a prediction of \(h_s\).

Now, focusing on Term~(I), the meta-learner leverages delayed optimistic FTRL/AdaHedge to \(M_t\), and updates its fusing weights by executing:
\begin{equation}\label{eq:d-opt-ogd:meta-weights}
w_{t+1}
=
\arg\min_{w\in\Delta_N}
\left\{
\left\langle w,M_{1:t-m}+\widetilde M_{t-m+1:t+1}\right\rangle
+
\eta^{\rm M}_{t+1}R_\Delta(w)
\right\}.
\end{equation}
where we defined $\widetilde M_t\!=\!X_t^\top \widetilde h_t$. Here we used an entropic regularizer \(R_\Delta(w)\!=\!\sum_{i=1}^N w_i\log w_i\), and the DUB regularization \(\eta^{\rm M}_{t+1}\!=\!\eta_{t+1}(\alpha_{\rm M})\) from \cite{flaspohler2021online}, which employs the formula:
\begin{equation}\label{eq:DUB_tuned_regularizer_1}
\eta_{t+1}(\alpha)
=
\frac{2}{\alpha}\max_{j\le t-m-1}a_{j-m+1:j}
+
\frac{1}{\alpha}
\sqrt{\sum_{r=1}^{t-m}(a_r^2+2\alpha b_r)} .
\end{equation}
Parameters \(a_t,b_t\!\ge\!0\) and \(\alpha\!>\!0\) are problem-specific, and for our meta-learning are defined as:
\begin{align}
&a_t^{\rm M}=2\|\bar e_t(M)\|_\infty,\quad
b_t^{\rm M}=\operatorname{huber}(\|\bar e_t(M)\|_\infty,\|\bar n_t(M)\|_\infty),
\quad
\alpha_{\rm M}=\log N, \quad \text{where} \label{eq:param_meta_DUB} \\
&\bar e_t(M):=M_{t-m:t}-\widetilde M_{t-m:t},
\quad
\bar n_t(M):=
M_{t-m}+\widetilde M_{t-m+1:t+1}-\widetilde M_{t-m:t}, \quad \text{and} \notag \\
&\operatorname{huber}(x,y)
:=
\frac12x^2-\frac12(x-y)_+^2
\le
\min\left\{\frac12x^2,xy\right\}.\notag
\end{align}
There are few interesting remarks here. First, the meta-learner operates on linear losses, so \(w_{t+1}\) itself does not need to be predicted. Second, before choosing \(w_{t+1}\), the meta-learner already knows \(X_{t+1}\) and hence the only unknown quantity is the future feedback \(h_{t+1}\). Finally, observe that we need to predict only the gradient vectors, not matrices \(X_t\) (see Lemma~\ref{lem:meta_bound} in Appendix).

Turning our attention to Term~(II), each base learner \(i\) is tuned to a path-length guess \(\rho_i\), using the dyadic grid of possible values, as in \cite{zhang2018adaptive}. However, here the base learners need also to handle the feedback delay and include the predictions. To enable this, we rely again on the toolbox of
\cite{flaspohler2021online} which we modify for dynamic regret. This extension involves a non-trivial analysis due to the interaction of the prediction errors with the comparator variation (see Theorem \ref{thm:DynamicOOGD} and Corol. \ref{cor:termII_dub_final} in Appendix); yet, interestingly, its implementation requires minimal changes. Namely, each base learner runs:
\begin{align}\label{eq:d-opt-ogd:base_learner}
&x_{t+1}^{(i)} =\arg\min_{x\in\mathcal X}
\left\{\left\langle h_{t-m}+\widetilde h_{t-m+1:t+1}-\widetilde h_{t-m:t}, x \right\rangle +
\frac{\eta_{t+1}^i}{2}\|x-x_t^{(i)}\|_2^2
\right\}, \\
&\eta^i_{t+1}\!=\!\eta_{t+1}(\alpha_i) \  \text{with} \ a_t^{i}\!=\!D\|\bar e_t(h)\|_2,\ \
b_t^{i}\!=\!\operatorname{huber}(\|\bar e_t(h)\|_2,\|\bar n_t(h)\|_2),
\ \
\alpha_i=D(D+2\rho_i).\label{eq:param_base_DUB}
\end{align}

\begin{algorithm}[t]
\caption{\small Universal Meta-Algorithm for Delayed Optimistic OCO}
\label{alg:meta-delayed-ooco}
\small
\begin{algorithmic}[1]
\Require Grid of path-length scales \(\{\rho_i\}_{i=1}^{N}\), initial weights \(w_m=\mathbf 1/N\).
\Require Constraint parameters \(V\), \(\Phi\), and \(\lambda\).
\State Activate \(N\) base learners by running Algorithm~\ref{alg:base-delayed-oogd}
with scale \(\rho_i\), for each \(i\in[N]\).
\For{\(t=m,\ldots,T\)}
    \State Receive \(x_t^{(i)}\) from each base learner and set
    \(X_t=[x_t^{(1)},\ldots,x_t^{(N)}]\).
    \State Play the convex combination
    \(
        x_t = X_t w_t .
    \)
    \State Observe the delayed feedback \(h_{t-m}\) and the hints \(\widetilde h_{t-m+1}^{t+1}\).
    \State Send \(h_{t-m}\) and \(\widetilde h_{t-m+1:t+1}\) to all base learners.
    \State Receive the next predictions \(x_{t+1}^{(i)}\) and set
    \(X_{t+1}=[x_{t+1}^{(1)},\ldots,x_{t+1}^{(N)}]\).
    \State Define the delayed meta-loss and optimistic hints
    \[
        M_{t-m}=X_{t-m}^{\top}h_{t-m},
        \qquad
        \widetilde M_s=X_s^\top \widetilde h_s,
        \quad s=t-m+1,\ldots,t+1 .
    \]
    \State Compute \(\eta^{\rm M}_{t+1}\) using
    \eqref{eq:DUB_tuned_regularizer_1} with \eqref{eq:param_meta_DUB}.
    \State Update \(w_{t+1}\) using \eqref{eq:d-opt-ogd:meta-weights}.
\EndFor
\end{algorithmic}
\end{algorithm}


\begin{algorithm}[t]
\caption{\small Delayed Optimistic OGD Base Learner}
\label{alg:base-delayed-oogd}
\small
\begin{algorithmic}[1]
\Require Path-length scale \(\rho\), initial points \(x_0,\ldots,x_m\).
\Require Constraint parameters \(V\), \(\Phi\), and \(\lambda\).
\For{\(t=m,\ldots,T\)}
    \State Submit \(x_t^{(i)}\) to the meta-algorithm.
    \State Receive the delayed feedback \(h_{t-m}\) and the hints
    \(\widetilde h_{t-m+1:t+1}\).
    \State Compute \(\eta_{t+1}^{\rho}\) using
    \eqref{eq:DUB_tuned_regularizer_1} with \eqref{eq:param_base_DUB}.
    \State Update \(x_{t+1}^{(i)}\) using \eqref{eq:d-opt-ogd:base_learner}.
\EndFor
\end{algorithmic}
\end{algorithm}

Alg. \ref{alg:meta-delayed-ooco} summarizes the execution of this meta-learning framework, and the next theorem its regret.
\begin{theorem}[Delayed optimistic OGD]
\label{thm:delayed_optimistic_oco}
Let \(h_t\) be the delayed gradient feedback and let
\(\tilde h_t\) be its prediction. For any comparator sequence \(u_{1}^T\in \mathcal U_T\), Algorithm \ref{alg:meta-delayed-ooco} ensures universal dynamic regret:
\[
\mathcal R_T(L)
=
\mathcal O\!\left(
(m+1)^{3/2}
\left(
1+\sqrt{1+\log N}
+
\sqrt{1+P_T}
\right)
\sqrt{
\sum_{t=1}^{T}\|h_t-\tilde h_t\|^2
}
\right).
\]
\end{theorem}

{\emph{Proof outline.} By convexity, the delayed loss regret is upper bounded by the delayed linearized regret,
\(\mathcal R_T(L)\le \mathcal R_T(\ell)\!=\!\text{Term (I)} \!+\! \text{Term(II)}.\) Term~(I) is the meta-regret of Alg.~\ref{alg:meta-delayed-ooco}, and is controlled by the delayed optimistic FTRL/AdaHedge update of \cite[Theorem~12]{flaspohler2021online} with entropic regularizer, see Lemma~\ref{lem:meta_bound}. Term~(II) is the delayed dynamic regret of the best-tuned base learner, Alg.~\ref{alg:base-delayed-oogd}, and is controlled by Theorem~\ref{thm:DynamicOOGD} and Corollary~\ref{cor:termII_dub_final}. Finally, the path-length grid, following the meta-learning idea of \cite{zhang2018adaptive}, ensures one base learner is tuned to $P_T$ up to constants.

\textbf{Discussion.} The factor \((m\!+\!1)^{3/2}\) should be interpreted as the cost of maintaining a
prediction-adaptive bound under delayed feedback. If predictions are ignored
and a uniform gradient bound is available, one can instead use
fixed worst-case tuning and obtain a delayed dynamic-regret bound with linear
dependence on \(m\!+\!1\). We refer to Appendix~\ref{app:proof_Term(I)}
for a more detailed discussion of this dependence. However, such a worst-case tuning would lose the benefit of accurate predictions and is less suitable for the constrained setting, where the surrogate gradients may not admit a useful uniform Lipschitz bound.  Under worst-case predictions accuracy, Theorem \ref{thm:delayed_optimistic_oco} guarantees regret $\mathcal O\big(m^{3/2}\sqrt{P_T}
\sqrt{\sum_{t=1}^{T}\|h_t\|^2},
\big)$, which matches the lower bound of \cite{zhang2018adaptive} for $m\!=\!1$, and the delay dependency (equivalently, memory length) of \cite{anava2015online} (static regret; no predictions). Interestingly, under perfect predictions we obtain regret $\mathcal R_T(L)=\mathcal O(1)$. The bound holds for any value of $m$ and recovers (up to constants) the respective no-memory bound \(\mathcal{O}\!(\sqrt{(1+P_T)E_T(f)})\) of \cite{mhaisen2025dynamic} in the known $P_T$ case, where $E_T(f)$ is the accumulated prediction error.  

Technically, the main difference from \cite{flaspohler2021online} is that, under a dynamic benchmark, prediction errors interact with the movement of comparator. We control this interaction using an auxiliary Young-type argument and show it does not change the desired prediction-adaptive order, except through the path-length dependence. We kindly refer the reader to Appendix \ref{sec:appendix-theorem1} for the derivations and discussion. From a computational perspective, we use multiple base learners to adapt to unknown $P_T$. Still, unlike previous delayed OCO frameworks e.g., \cite{joulani2013online}, we use only one copy of the algorithm for each $P_T$ value, instead of \(m+1\) parallel copies that would have been trained only on a fraction of data.

%% file: Neurips/COCO.tex
\section{Handling Time-average Adversarial Constraints}\label{sec:COCO}


This section extends the above analysis to handle time-average adversarially-varying constraints, i.e., to \emph{constrained} optimistic OCO with delayed feedback. To facilitate presentation, we use the shifted notation from Section~\ref{sec:OCO}: the feedback received at round \(t\) corresponds to decision \(x_{t-m}\). COCO methods perform OCO over a time-varying Lagrangian-type surrogate using either a primal--dual method, e.g., \cite{lamperski}, or a penalty method where the accumulated constraint violation, \(Q_t\), is
penalized directly in a scalarized surrogate loss \( L_t(x)\) \cite{lekeufack2024optimistic,sinha2024optimal, sinha2025beyond, pmlr-v97-liakopoulos19a}. We follow the latter approach and define:
\begin{equation}
L_t(x_{t-m})
=
V f_t(x_{t-m})
+
\Phi'\!\bigl(Q_{t-m-1}\bigr)\,
g_t^+(x_{t-m}),
\qquad
Q_t=Q_{t-1}+g_t^+(x_{t-m}),
\label{eq:delay-surrogate}
\end{equation}
where \(V\!>\!0\) is a trade-off parameter, and \(\Phi\) an increasing convex penalty function. Equivalently, in the decision-time notation of Sec.~\ref{sec:OCO}, one sets
\(s\!=\!t\!-\!m\), so that the feedback associated with \(x_s\) is the gradient of the surrogate loss revealed at \(s+m\). The penalty coefficient is evaluated at \(Q_{t-m-1}\), which is available when \(x_{t-m}\) is decided. This avoids the need to predict the future queue evolution when forming the gradient prediction. As a result, the prediction error of the surrogate gradient separates naturally into the prediction error of the objective and that of the constraint violation.

We define the prediction errors at the level of the delayed gradients. For every \(s\in[T]\), set
\begin{align*}
&\Delta_s^f
:=
\nabla f_{s+m}(x_s)
-
\nabla \tilde f_{s+m}(\tilde{x}_s),
\qquad
E_T(f):=\sum_{s=1}^{T}\|\Delta_s^f\|_2^2,
\end{align*}
The term \(E_T(g^+)\) is defined analogously. At \(t\), all past decisions are known, thus suffices to predict only \(\widetilde x_{t+1}\). We now recall the decomposition connecting the surrogate regret to $\mathcal{R}_T(f)$ and $Q_T$.
\begin{lemma}[adapted from \cite{Abdullah_Iosifidis_Elayoubi_Chahed_2026,supantha2025universal}]
\label{lem:delay-prediction-forward-decomposition}
For any differentiable, convex and increasing penalty \(\Phi\), it is:
\begin{align}
\Phi(Q_T)-\Phi(Q_{m-1})
+
V\mathcal{R}_T(f)
\le
\mathcal{R}_T(L)
+
G(m+1)\Phi'(Q_T),
\label{eq:delay-decomposition-main}
\end{align}
where the dynamic regret is computed with respect to comparator sequences \(u_1^T\in\mathcal U_T\).
\end{lemma}

\begin{remark}\normalfont
    The queue is maintained globally using the actually played trajectory. After the learner plays \(x_t=X_t w_t\), the queue is updated from the realized constraint violation, and the corresponding surrogate feedback \(h_t\) is constructed and fed to all base learners and to the meta-learner. Thus, the base learners do not maintain separate counterfactual queues.
\end{remark}

\subsection{Unknown Path Length: Regret and CCV Bounds}

We first consider the case where \(P_T\) is unknown. We apply the meta-base delayed optimistic algorithm developed in Section~\ref{sec:OCO} to the surrogate loss defined in \eqref{eq:delay-surrogate}.

\begin{theorem}[Unknown-path-length bounds with CCV-oriented tuning]
\label{thm:unknown-path-quadratic-ccv}
Let \(u_1^T\in\mathcal U_T\) be any feasible comparator sequence with path length \(P_T\). Using a quadratic penalty $\Phi(Q)=Q^2$, and
\[
V
:=
4(m+1)^{3/2}
\left(1+\sqrt{1+\log N}\right)^2
\sqrt{E_T(g^+)}
+
2G^2(m+1)
+
Q_{m-1}^2 .
\]
Then, under the delayed optimistic meta-base algorithm \ref{alg:meta-delayed-ooco} the following bounds hold

\begin{equation}\label{eq:unknown-path-ccv-regret-final}
\mathcal{R}_T(f)
\le
\mathcal O\!\left(
A\sqrt{E_T(f)}
+
(m+1)^{3/2}(1+P_T)\sqrt{E_T(g^+)}
+
(m+1)
\right), \qquad \text{and} \notag
\end{equation}
\begin{equation}
Q_T\!\le\!
\mathcal O\!\left(\!
A\sqrt{T}\,E_T(g^+)^{1/4}\!+\!
(m\!+\!1)^{\frac{5}{4}}\sqrt{T}
\!\right) \ \text{with} \ 
A\!=\!
\mathcal O\!\left(\!
(m\!+\!1)^{\frac{3}{2}}
\left(
\!1\!+\!\sqrt{1\!+\!\log N}+\sqrt{1\!+\!P_T}
\right)\!
\right)\notag
\end{equation}

We note that \(P_T\) appearing in the bound on \(Q_T\) can be taken as the smallest feasible path length. In particular, it can be zero whenever there exists a fixed decision that satisfies all constraints.
\end{theorem}


\textbf{Discussion.} Theorem~\ref{thm:unknown-path-quadratic-ccv} gives an optimistic delayed-COCO guarantee under a universal dynamic benchmark. Under worst-case predictions, we obtain an
\(\mathcal{O}((1+P_T)\sqrt{T})\) regret bound and an
\(\mathcal{O}\!\left(\max\{T^{3/4},\sqrt{(1+P_T)T}\}\right)\) CCV bound, recovering the best known rates of \cite{supantha2025universal} when \(m=0\). Under perfect predictions, our bounds reduce to \(\mathcal{O}(1)\) regret and \(\mathcal{O}(\sqrt{T})\) CCV. We note that \cite{yi2021regret} achieves \(\mathcal{O}(\sqrt{(1+P_T)T})\) regret and \(\mathcal{O}(\sqrt{T})\) CCV for time-invariant constraints. Our perfect-prediction regime recovers the same CCV order while handling time-varying constraints. In the known $P_T$ setting, we get better bounds as shown later. As discussed in Appendix~\ref{app:proof_Term(I)}, the extra \(\sqrt{m+1}\) factor in the simplified delayed-regret bound can be absorbed by the meta-complexity or path-length terms when \(\log N\) or \(P_T\) dominates \(m\), leading to linear dependence on the delay. Finally, since \(V\) depends on \(E_T(g^+)\), we use a doubling trick to learn this quantity online, introducing at most an \(\mathcal{O}(\log T)\) overhead, as is standard in optimistic COCO, see, e.g., \cite{lekeufack2024optimistic,Abdullah_Iosifidis_Elayoubi_Chahed_2026}.


\begin{remark}[Penalty choice and path-length dependence]\normalfont
In COCO, the surrogate loss must do two things: track the comparator for regret and penalize constraint violations to control the queue. This coupling between tracking and feasibility is what leads to a higher dependency on \(P_T\). If \(P_T\) is known, the penalty parameter can be tuned to the true scale of comparator motion, allowing balancing regret and queue control. When $P_T$ is unknown, this tuning is not available. Moreover, the usual meta-learning approach cannot be applied directly to the penalty parameter, since different penalty choices generate different surrogate loss sequences. Thus, one must use a single surrogate construction that works uniformly; this leads to a more conservative trade-off.

Also, the \(T^{3/4}\)-type term in the CCV bound comes from the same limitation. We use the quadratic penalty \(\Phi(Q)=Q^2\), which is the smallest polynomial penalty that still yields sublinear queue growth. Stronger penalties, such as \(\Phi(Q)=Q^\beta\) with \(\beta>2\), or exponential penalties, may improve the CCV exponent, but they also amplify the path-length contribution in the surrogate regret bound. Thus, improving CCV in this way comes at the cost of worse dependence on \(P_T\). We study next the known-path-length regime, where this trade-off can be tuned more carefully. 
\end{remark}

\subsection{Known Path Length}

When \(P_T\) is known we can tune the surrogate loss directly to the scale of the comparator motion via the penalty \(\Phi\). Hence, we can direcly run only one base learner, Algorithm~\ref{alg:base-delayed-oogd}, with \(\rho=P_T\).

\begin{theorem}[Known-path-length case with exponential penalty]
\label{thm:known-PT}
For any $u_{1}^{T}\in \mathcal{U}_T$ with path length \(P_T\) to be known. Choose \( \rho=P_T, V=1\), and \(\Phi(Q)=e^{\lambda Q}-1\), with:
\[
\lambda
=
\frac{1}{
2\left(
4(m+1)^{3/2}
\left(
D+\sqrt{D(D+P_T)}
\right)
\sqrt{E_T(g^+)}
+
G(m+1)
\right)
}.
\]
Feeding these parameters into Algorithm~\ref{alg:base-delayed-oogd}, the following bounds hold:
\begin{equation}
\label{eq:known-PT-exp-reg-clean}
\mathcal{R}_T(f)
\le
\mathcal{O}\!\left(
(m+1)^{3/2}\sqrt{(1+P_T)E_T(f)}
\right), \qquad \text{and}
\end{equation}
\begin{equation}
\label{eq:known-PT-exp-ccv-clean}
Q_T
\le
\mathcal{O}\!\left(
\left(
(m+1)^{3/2}\sqrt{(1+P_T)E_T(g^+)}
+
G(m+1)
\right)
\log T
\right).
\end{equation}

\end{theorem}

\textbf{Discussion.} Theorem~\ref{thm:known-PT} gives prediction-dependent bounds for the known $P_T$ case, without using meta-learning. We first compare with dynamic optimistic COCO works with predictions. When \(m=0\), we recover the no-memory optimistic COCO bounds of \cite{lekeufack2024optimistic}:
$\mathcal R_T(f)
=
\mathcal O\!\big(\sqrt{P_TE_T(f)}\big),\: Q_T
=
\mathcal O\!\big(
\bigl(\sqrt{P_TE_T(g^+)}+G\bigr)\log T
\big).
$
Thus, our result extends their prediction-dependent guarantees to delayed and memory-dependent settings. The closest static optimistic memory-dependent works are \cite{Abdullah_Iosifidis_Elayoubi_Chahed_2026,mhaisen2024optimistic}. These works rely on separability of the memory terms, which facilitates isolating the contribution of each decision, but leads to a more restrictive benchmark in constrained problems and restricted class of loss functions. Our approach avoids this assumption and recovers the same type of prediction-dependent bounds while preserving the original joint feasibility benchmark. Finally, compared to \cite{liu2023online}, which studies the static memory-constrained problem without predictions, our bounds improve over their \(\tilde{\mathcal O}(T^{2/3})\) regret and CCV rates.

\begin{remark}\normalfont
If \(P_T\) is unknown but measurable online, a standard doubling trick recovers
the bounds of this subsection. The same applies to tuning \(\lambda\), since
\(E_T(g^+)\) is observable online.
\end{remark}

%% file: Neurips/memory.tex
\section{Memory as Delayed Feedback}
\label{sec:memory_delay}

We now transfer the delayed optimistic dynamic regret guarantees to the memory problems. In this setting, the loss and constraint functions depend on a block of past decisions, \(f_t(x_{t-m}^{t})\) and \(g_t(x_{t-m}^{t})\). Thus, \(x_s\) affects all losses whose memory
window contains it, i.e., all \(t\) such that \(t-m\le s\le t\). Hence, its full
first-order contribution is revealed only after these losses are observed,
which induces a feedback delay of at most \(m\). Note that we redefine here the learning functions, $L_t$, and the horizon ($T_m$ instead of $T$) (further explanations about notation conventions, please see Appendix Sec. \ref{sec:app_section5}).

The reduction is performed at the regret level. The memory losses are not
assumed to be linear. Rather, we use the standard OCO linearization argument:
applying the first-order convexity inequality to each memory surrogate loss
\(L_t\) around the played block \(x_{t-m}^{t}\) gives an upper bound in terms of
partial gradients. More precisely, when \(L_t\) is nondifferentiable, we select a joint subgradient
\[
G_t=(G_{t,t-m},\ldots,G_{t,t})
\in
\partial L_t(x_{t-m}^{t}),
\]
and interpret \(\nabla_{x_s}L_t(x_{t-m}^{t})\) below as the coordinate block
\(G_{t,s}\). Thus, all coordinate blocks associated with a given \(L_t\) are
taken from the same joint subgradient, the same convention applies to predicted functions. For each decision index \(s\), define
\begin{equation}
I_s:=\{t\in [T_m]:\ t-m\le s\le t\},
\qquad
h_s:=\sum_{t\in I_s}\nabla_{x_s}L_t(x_{t-m}^t).
\label{eq:auxiliary-to-lemma-2}
\end{equation}
Here, \(I_s\) is the set of losses whose memory window contains \(x_s\), and
\(h_s\) is the corresponding \emph{credit gradient}. It induces the auxiliary
linearized loss \(\ell_s(z):=\langle h_s,z\rangle\). Since \(I_s\subseteq\{s,\ldots,s+m\}\), the vector \(h_s\) is fully revealed no later than round \(s+m\).

\begin{lemma}[Memory-to-delay linearization]
\label{lem:memory_to_delay}
Let \(L_t:\mathcal X^{m+1}\!\mapsto\!\mathbb R\) be jointly convex
\(\forall t\in[T_m]\), and define \(I_s\), \(h_s\) as in
\eqref{eq:auxiliary-to-lemma-2}, using the joint-subgradient convention above whenever \(L_t\) is non-differentiable. Then, for any comparator sequence
\(u_1^T\in\mathcal U_T\), it holds that
\[
\sum_{t=m}^{T}
L_t(x_{t-m}^{t})-L_t(u_{t-m}^{t})
\le
\sum_{s=1}^{T}
\langle h_s,x_s-u_s\rangle .
\]
Moreover, \(h_s\) is revealed with delay at most \(m\).
\end{lemma}

The following corollary transfers the delayed guarantees to the memory setting.

\begin{corollary}[Transfer from delayed OCO to memory]
\label{coro:memory_to_delay}
Any regret guarantee for delayed OCO with delay at most \(m\), applied to the
auxiliary linearized losses \(\ell_s(z)=\langle h_s,z\rangle\), yields the
corresponding guarantee for the memory-dependent regret. In particular,
Theorems~\ref{thm:delayed_optimistic_oco},
\ref{thm:unknown-path-quadratic-ccv}, and \ref{thm:known-PT} apply to the
memory-dependent problem after replacing the delayed gradients by the credit
gradients \(h_s\). The corresponding
objective and constraint prediction errors induced by the credit gradients are specified below.
\end{corollary}

 In the delayed case without memory, the hint predicts one delayed gradient, e.g.,
\(\nabla f_{s+m}(x_s)\), and constructing the next hint only requires a
one-step proxy decision. In the memory case, the hint must predict the whole
credit gradient \(h_s\). Thus, one may set
\begin{equation}
\widetilde h_s
:=
\sum_{t\in I_s}
\nabla_{x_s}\widetilde L_t(\widetilde x_{t-m}^{t}).
\label{eq:memory-predicted-credit-gradient}
\end{equation}
Consequently, constructing \(\widetilde h_{s+1}\) may require predicting the
future memory block up to \(\widetilde x_{s+m+1}\), not only the next decision
\(\widetilde x_{s+1}\). Thus, the hints used by
Algorithm~\ref{alg:meta-delayed-ooco} when \(P_T\) is unknown, or directly by
Algorithm~\ref{alg:base-delayed-oogd} when \(P_T\) is known, are 
\[
\widetilde h_{s-m+1:s+1}
=\sum_{\tau=s-m+1}^{s+1}
\sum_{t\in I_\tau}
\nabla_{x_\tau}\widetilde L_t(\widetilde x_{t-m}^{t}).
\]

\begin{remark}\label{Rem:memory_errors}\normalfont
In the constrained case, we use the surrogate loss
\begin{equation}
L_t(x_{t-m}^{t})
=
V f_t(x_{t-m}^{t})
+
\Phi'(Q_{t-m-1})\,g_t^+(x_{t-m}^{t}),
\qquad
Q_t=Q_{t-1}+g_t^+(x_{t-m}^{t}).
\label{eq:memory-surrogate-final}
\end{equation}
The regret decomposition Lemma of
\cite{Abdullah_Iosifidis_Elayoubi_Chahed_2026} applies to this surrogate too
for \(u_1^T\in\mathcal U_T\). Thus, once the dynamic regret of \(L_t\) is
controlled via Corollary~\ref{coro:memory_to_delay}, the objective-regret and
CCV bounds follow as in the delayed case. Equivalently, the prediction errors for the final COCO-M bounds are defined at
the level of the objective and constraint credit gradients, 
\(E_T^{\mathrm{mem}}(f)\) and \(E_T^{\mathrm{mem}}(g^+)\), where

\begin{align}
E_T^{\mathrm{mem}}(f)
&:=
\sum_{s=1}^{T}
\left\|
\sum_{t\in I_s}
\left[
\nabla_{x_s}f_t(x_{t-m}^{t})
-
\nabla_{x_s}\widetilde f_t(\widetilde x_{t-m}^{t})
\right]
\right\|^2 ,
\label{eq:memory-error-f}
\\
E_T^{\mathrm{mem}}(g^+)
&:=
\sum_{s=1}^{T}
\max_{r\in I_s}
\left\|
\sum_{\substack{t\in I_s\\ t\ge r}}
\left[
\nabla_{x_s}g_t^+(x_{t-m}^{t})
-
\nabla_{x_s}\widetilde g_t^+(\widetilde x_{t-m}^{t})
\right]
\right\|^2 .
\label{eq:memory-error-g}
\end{align}
The latter quantity provides a penalty-independent upper envelope for
the normalized constraint credit-gradient prediction error, as shown
in Appendix~\ref{sec:app_section5}.
\end{remark}

\textbf{Discussion.} Prior optimistic approaches to OCO with memory
\cite{mhaisen2024optimistic,Abdullah_Iosifidis_Elayoubi_Chahed_2026}, 
isolate the contribution of each decision at the level of loss functions, which requires separability across the memory variables. 
Our reduction keeps the memory-dependent loss in its original jointly convex form and linearizes only at the regret level. The price is that optimistic hints must predict credit gradients rather than single gradients. This may require predicting future memory blocks, but avoids imposing component-wise feasibility on separable parts of the constraint, and preserves the original joint feasibility benchmark. For the constraints, different memory contributions carry different queue-dependent penalty coefficients. We therefore use a penalty-independent envelope of the aggregated prediction errors, which parallels the separable setting \cite{Abdullah_Iosifidis_Elayoubi_Chahed_2026} while controlling the
corresponding weighted error. It vanishes under perfect predictions
and preserves an \(O(\sqrt{T})\) worst-case dependence, up to
memory-dependent factors.

%% file: Neurips/Conclusion.tex
\section{Conclusions and Limitations} \label{sec:conclusion}

The proposed framework contributes in closing a gap in COCO with memory, and NSC with time-average budget constraints, and can be used to tackle important problems in ML, operations research, control of cyber-physical systems, and so on. Aiming for universal dynamic regret ensures the learning is against the most demanding comparators; while the inclusion of optimism enables the bounds to improve, reaching path-length-oblivious regret \(\mathcal R_T\!=\!\mathcal O(1)\), together with \(Q_T\!=\!\mathcal O(\sqrt{T})\) when \(P_T\) is unknown, and \(Q_T\!=\!\mathcal O(\log T)\) when known or observable at runtime. 

There are few limitations which call for follow up work. First, the framework requires predicting, apart from future functions, also some future actions. This cyclic dependency is implicit in almost all optimistic learning works (unless the losses are linear); still, it is worth investigating whether it can be avoided, see discussion in \cite{scroccaro2023adaptive}. Secondly, the constraint violation does not scale down to $\mathcal O(1)$ under perfect predictions. While this matches the best-known optimistic COCO bounds (without memory), it is certainly exciting to explore further improvements. Finally, expanding the set of eligible benchmarks in COCO is a key challenge; our work contributes in this direction but the question of approaching closer to horizon-long feasibility remains open.

%% file: Neurips/appendix.tex
\setcounter{section}{0}
\renewcommand{\thesection}{\Alph{section}}
\renewcommand{\thesubsection}{\Alph{section}.\arabic{subsection}}

The structure of this appendix is as follows.
\begin{enumerate}
\item Section~\ref{sec:appendix-related} provides a comprehensive discussion of related work.

\item Section~\ref{sec:appendix-theorem1} provides the proof of Theorem~\ref{thm:delayed_optimistic_oco}. This requires deriving a dynamic-regret bound for our delayed optimistic OCO algorithm; the section includes both the auxiliary results and their proofs.

\item Section~\ref{sec:app_sec4} provides the proofs of Theorem~\ref{thm:unknown-path-quadratic-ccv} and Theorem~\ref{thm:known-PT}, respectively in Sections~\ref{subsec:app_proof_theorem2} and~\ref{subsec:proof_theorem3}.

\item Section~\ref{sec:app_section5} provides the proofs of the memory-to-delay equivalence results stated in the main paper, namely Lemma~\ref{lem:memory_to_delay} and Corollary~\ref{coro:memory_to_delay}.

\end{enumerate}

\section{Discussion of Related Work} \label{sec:appendix-related}

This section provides a detailed discussion of related works. Table \ref{tab:related-work} presents a concise high-level comparison with selected prior results. The proposed method is at the nexus of different OCO flavors, namely OCO with time-average adversarially-changing constraints, OCO with memory, OCO with delay and OCO with predictions. Therefore, we review below most relevant works.

\input{Neurips/relatedwork}

\section{Proofs of Section \ref{sec:OCO}} \label{sec:appendix-theorem1}

\subsection{Proof of Theorem \ref{thm:delayed_optimistic_oco}}

We provide the proof of Theorem \ref{thm:delayed_optimistic_oco}, which we restate here for the readers convenience.
\begin{mdframed}
Let \(h_t\) be the delayed gradient feedback and let
\(\tilde h_t\) be its prediction. For any comparator sequence \(u_{1}^T\in \mathcal U_T\), Algorithm \ref{alg:meta-delayed-ooco} ensures dynamic regret:
\[
\mathcal R_T(L) = \mathcal O\!\left( (m+1)^{3/2} \left( 1+\sqrt{1+\log N}+\sqrt{1+P_T}\right)\sqrt{\sum_{t=1}^{T}\|h_t-\tilde h_t\|^2}
\right).
\]
\end{mdframed}

The proof of this result comprises two stages. First, we bound \textnormal{Term (II)} and then \textnormal{Term (I)}. For the former, we introduce an optimistic dynamic regret variant of OGD that uses adaptive regularization and can handle delays. For \textnormal{Term (I)}, we adapt the static regret delayed optimistic algorithm of \cite{flaspohler2021online}.

\subsubsection{Bounding Term(II)}

We first analyze the undelayed optimistic problem. This step is useful because the delayed update will later be recovered through the delay-as-optimism reduction. In the dynamic setting, the main challenge is the movement of the comparator. We therefore use the projected one-step geometry, which yields a clean telescoping argument and exposes the additional interaction between the prediction error and the comparator variation. We introduce the following algorithm and respective bound:

\begin{theorem}[Optimistic dynamic OGD]\label{thm:DynamicOOGD}
Let \((\eta_t)_{t\ge 1}\) be a nondecreasing positive sequence. Let
\(L_t:\mathcal X\to\mathbb R\) be convex losses, and let
\(h_t=\nabla L_t(x_t)\). In this theorem, feedback is non-delayed ( after playing \(x_t\) the learner observes \(h_t\)). consider the projected optimistic one-step update
\[
x_{t+1}
=
\Pi_X\!\left(
x_t-\frac{1}{\eta_{t+1}}\bigl(h_t+\tilde h_{t+1}-\tilde h_t\bigr)
\right),
\qquad x_1=\Pi_X\!\left(x_0-\frac{1}{\eta_1}\tilde h_1\right),
\]

we have that for any comparator sequence $u_1,\dots,u_T\in \mathcal{U}_T$, the regret satisfies:
\begin{equation}\label{eq:optimistic_Dynamic_OGD}
    \mathcal{R}_T(L)\le
\eta_{T+1}\Bigl(D^2+\tfrac32 D\,P_T\Bigr)
+
\sum_{t=1}^T
\min\!\left(
D\|e_t\|_2,\;
\frac{1}{\eta_{t+1}}\operatorname{huber}\!\bigl(\|e_t\|_2,\|n_t\|_2\bigr)
\right)
-\sum_{t=1}^T \langle e_t,\delta_t\rangle, 
\end{equation}
where
\(
e_t:=h_t-\tilde h_t\), \(n_t:=h_t+\tilde h_{t+1}-\tilde h_t\), and \(\delta_t:=u_t-u_{t+1}.
\)
\end{theorem}

Theorem~\ref{thm:DynamicOOGD} will serve as a technical bridge with the delayed analysis. Related optimistic dynamic-regret guarantees have been studied before; see, for instance, \cite{jadbabaie2015online} and, more recently \cite{scroccaro2023adaptive} and \cite{mhaisen2025dynamic}. However, these results are developed for the standard non-delayed feedback model and cannot be used here. The purpose of Theorem~\ref{thm:DynamicOOGD} is to derive a non-delayed optimistic dynamic-regret bound in a form compatible with the delayed-upper-bound (DUB) argument, both in the \(m\)-step delayed setting of \cite{flaspohler2021online} and in the one-step delayed adaptive-tuning argument of \cite{orabona2019modern}. After applying the delay-as-optimism reduction, this structure will allow us to choose the regularization adaptively from observed quantities, avoiding over-regularization and preserving the benefit of accurate predictions. The additional term compared with the static-benchmark case of \cite{flaspohler2021online} is
\(
-\sum_{t=1}^{T}\langle e_t,\delta_t\rangle,
\)
which couples the prediction error with the movement of the comparator.

We now pass from the undelayed optimistic bound to the delayed setting. By Lemma~2 of \cite{flaspohler2021online}, delayed feedback can be viewed as optimism with a particular choice of ``bad'' hint. In our notation, this hint is
\[
\tilde h_{t+1}^{\mathrm{bad}}
:=
\tilde h_{t-m+1:t+1}-h_{t-m+1:t}.
\]
 Applying this representation to the bound in \eqref{eq:optimistic_Dynamic_OGD}, we obtain the same form of guarantee after replacing \(e_t\) and \(n_t\) by
\[
\bar e_t
:=
h_{t-m:t}-\tilde h_{t-m:t},
\qquad
\bar n_t
:=
h_{t-m}+\tilde h_{t-m+1:t+1}-\tilde h_{t-m:t}.
\]
The algebraic verification of this substitution is deferred to Appendix~\ref{app:delay as optimism}, using Corollary~\ref{cor:dynamic_delay_as_optimisim} presented there.

The genuinely new term in the dynamic optimistic analysis is
\[
-\sum_{t=1}^{T}\langle \bar e_t,\delta_t\rangle,
\qquad
\delta_t:=u_t-u_{t+1}.
\]
It appears because the comparator used here changes from \(u_t\) to \(u_{t+1}\). In a standard non-optimistic analysis, this change only contributes through the path length. In the optimistic analysis, however, the update direction depends on predictions, so the movement of the comparator interacts with the prediction error
\(
\bar e_t.
\)
The term \(-\langle \bar e_t,\delta_t\rangle\) therefore measures whether the prediction error helps or hurts the tracking of the moving comparator. However this term is yet controlled and upper bounded by the desired bound that is of order $\sqrt{T}$ in the worst case and shrinks with prefect prediction and/or with static benchmark to be zero.

\begin{corollary}[DUB-tuned bound for Term~(II)]
\label{cor:termII_dub_final}
Let
\[
a_t:=D\|\bar e_t\|_2,
\qquad
b_t:=\operatorname{huber}(\|\bar e_t\|_2,\|\bar n_t\|_2),
\]
Assume that the base learner uses the DUB-tuned regularization
\begin{equation}\label{eq:DUB_tuned_regularizer}
\eta_{t+1}
=
\frac{2}{\alpha}\max_{j\le t-m-1} a_{j-m+1:j}
+
\frac{1}{\alpha}
\sqrt{\sum_{i=1}^{t-m}\bigl(a_i^2+2\alpha b_i\bigr)},
\quad
\text{ with }
\alpha:=D\bigl(D+2P_T\bigr).
\end{equation}
Then the delayed dynamic regret of the best-tuned base learner satisfies
\[
\textnormal{Term (II)}
\le
4(m+1)^{3/2}
\left(
D+\sqrt{D(D+P_T)}
\right)
\sqrt{
\sum_{t=1}^{T}\|h_t-\tilde h_t\|^2
}.
\]
\end{corollary}

Compared with \cite{wan2024non}, our method is adaptive and does not require an a priori Lipschitz bound on the gradients. The standard black-box reduction of \cite{joulani2013online} handles delays by running multiple delay-free copies of a base algorithm, but this replication can be computationally costly and each copy is trained only on a partial subsequence of the observations. Other works avoid replication and incorporate delays directly into the analysis. For example, \cite{joulani2016delay} derives adaptive FTRL/Mirror-Descent bounds through a single-instance reduction based on prediction drift. More recently, \cite{hsieh2022multi} develops delayed dual averaging with data- and delay-dependent guarantees, including an optimistic variant. These works, however, focus on static regret. Our contribution is complementary. Building on the optimistic delayed-OCO perspective of \cite{flaspohler2021online}, we develop delayed optimistic guarantees for a dynamic comparator sequence. This distinction is important because static-regret techniques do not directly control the path length of a moving comparator. We then use these dynamic delayed-OCO guarantees as a bridge to constrained OCO with memory through the credit-gradient reduction.

\subsubsection{Bounding Term(I)}\label{app:proof_Term(I)}

We now turn our focus to Term(I) and bound it using an optimistic delayed AdaHedge algorithm. Using \(x_t=X_t w_t\) and \(x_t^i=X_t e_i\), it follows that \( x_t-x_t^i=X_t(w_t-e_i)\), thus we can write:
{\[
\textnormal{Term (I)}
\le
\sum_{t=1}^{T}\langle M_t,\;w_t-e_i\rangle, \quad \textnormal{where} \: M_t:=X_t^\top h_t.
\]}

This is the meta-regret of the delayed expert-combination scheme. This part follows from the delayed optimistic OCO framework of \cite{flaspohler2021online} applied to the linear meta-losses \(M_t=X_t^\top h_t\).

\begin{lemma}[Bound on Term~(I)]\label{lem:meta_bound}
Under the assumptions above, and using the static delayed optimistic FTRL meta-learner of \cite[Theorem~12]{flaspohler2021online}, for every expert \(i\in[N]\), we have:
\begin{equation}
\label{eq:termI-epsilon-bound}
\text{Term (I)}
\le
8D(m+1)^\frac{3}{2}\left(
1
+
\sqrt{1+\log N}
\right)\sqrt{\sum_{t=1}^{T}\|h_t-\tilde h_t\|^2}.
\end{equation}
\end{lemma}

$\bullet$ Finally, the proof of Theorem \ref{thm:delayed_optimistic_oco} is obtained by recalling that:
\[
\mathcal R_T(L) \le \mathcal R_T(\ell) =\textnormal{Term (I)}+\textnormal{Term (II)},
\]
and hence the results follows by combining Lemma~\ref{lem:meta_bound} with
Corollary~\ref{cor:termII_dub_final}.

\paragraph{Discussion} The factor \((m+1)^{3/2}\) in Theorem~\ref{thm:delayed_optimistic_oco} should be understood as the price of obtaining a prediction-adaptive bound under delayed feedback. The factor \(m+1\) comes from the block prediction error
\(
\bar e_t=h_{t-m:t}-\tilde h_{t-m:t},
\)
whereas the additional \(\sqrt m\) comes from the delayed-upper-bound (DUB) argument. More precisely, it appears when controlling the unrevealed window term
\[
\max_{t\in[T]} a_{t-m:t-1}
\le
D\sqrt m
\sqrt{\sum_{t=1}^{T}\|\bar e_t\|^2}.
\]
This term is the cost of choosing the regularization adaptively from delayed observations, which avoids over-regularization and preserves the benefit of accurate predictions.

If predictions are ignored and a uniform gradient bound is available, one can instead use a fixed worst-case tuning and recover a delayed dynamic-regret bound with linear dependence on \(m+1\). Thus, the \((m+1)^{3/2}\) dependence is not intrinsic to delayed OCO itself, but to the adaptive optimistic tuning under delay.

We also note that the sharper bounds in \eqref{eq:Term(I)_exact_bound} and
\eqref{eq:Term(II)_exact_bound} retain more information than the simplified
\((m+1)^{3/2}\) statement. In particular, within the prediction-adaptive bound
itself, the additional \(\sqrt m\) factor in Term~(I) can be dominated by the
meta-complexity term when \(m=O(\log N)\), while the corresponding term in
Term~(II) can be dominated by the path-length term when \(P_T\gtrsim Dm\).

\begin{remark}
One may alternatively implement the adaptive DUB tuning of \cite[Theorem~12]{flaspohler2021online}, which yields the same theoretical order, but yields for better performance in practice. 
\end{remark}

\subsection{Auxiliary Results and Proofs for Theorem \ref{thm:delayed_optimistic_oco} and Section \ref{sec:OCO}}

This subsection provides the intermediate results that are required for the proof of the main result of Section \ref{sec:OCO}, namely of Theorem \ref{thm:delayed_optimistic_oco} that we proved in the previous subsection.

\subsubsection{Proof of Theorem \ref{thm:DynamicOOGD}}

We remind the reader that this theorem assumes no delays, and thus after committing $x_t$ at round $t$, the gradient $h_t=\nabla_{x_t}L_{t}(x_t)$ is revealed directly and there is no delay of $m$ steps. 

Recall that:
\[
e_t:=h_t-\tilde h_t,
\qquad
n_t:=h_t+\tilde h_{t+1}-\tilde h_t,
\qquad
v_t:=x_t-x_{t+1},
\qquad
\delta_t:=u_t-u_{t+1},
\]

\begin{proof}
Since each $L_t$ is convex and $h_t= \nabla_{x_t} L_t(x_t)$, we have
\[
L_t(x_t)-L_t(u_t)\le \langle h_t,x_t-u_t\rangle, \qquad \text{and thus} \qquad \mathcal{R}_T(L)
\le
\sum_{t=1}^T \langle h_t,x_t-u_t\rangle.
\]
\medskip

\noindent
\textbf{Step 1: Exact decomposition.}
For each $t\in\{1,\dots,T\}$, we have
\begin{equation}
\label{eq:exact-decomp}
\langle h_t,x_t-u_t\rangle
=
\langle e_t,v_t\rangle
+
\langle n_t,x_{t+1}-u_t\rangle
+
\langle \tilde h_t,x_t-u_t\rangle
-
\langle \tilde h_{t+1},x_{t+1}-u_t\rangle .
\end{equation}

Summing \eqref{eq:exact-decomp} over $t=1,\dots,T$ yields
\begin{align}
\sum_{t=1}^T \langle h_t,x_t-u_t\rangle
&=\sum_{t=1}^T \langle n_t,x_{t+1}-u_t\rangle
+
\sum_{t=1}^T \langle e_t,v_t\rangle
\notag\\
&+\sum_{t=1}^T\Big(\langle \tilde h_t,x_t-u_t\rangle
-\langle \tilde h_{t+1},x_{t+1}-u_t\rangle\Big).
\label{eq:sum1}
\end{align}

Now the last sum telescopes as follows:
\begin{align}
\sum_{t=1}^T
\Big(
\langle \tilde h_t,x_t-u_t\rangle
-
\langle \tilde h_{t+1},x_{t+1}-u_t\rangle
\Big)
=\;&
\langle \tilde h_1,x_1-u_1\rangle
+
\sum_{t=1}^{T}\langle \tilde h_{t+1},\delta_t\rangle ,
\label{eq:telescope}
\end{align}
where the last equality uses $\delta_T=0$ and $\tilde h_{T+1}=0$. Substituting \eqref{eq:telescope} into \eqref{eq:sum1}, we obtain:
\begin{equation}
\label{eq:sum2}
\sum_{t=1}^T \langle h_t,x_t-u_t\rangle
=
\sum_{t=1}^T \langle n_t,x_{t+1}-u_t\rangle
+
\sum_{t=1}^T \bigl(\langle e_t,v_t\rangle +\langle \tilde h_{t+1},\delta_t\rangle\bigr)
+
\langle \tilde h_1,x_1-u_1\rangle.
\end{equation}
Since $n_t=e_t+\tilde h_{t+1}$, we have:
\[
\langle e_t,v_t\rangle+\langle \tilde h_{t+1},\delta_t\rangle
=
\langle e_t,v_t-\delta_t\rangle+\langle n_t,\delta_t\rangle.
\]
Hence, it is:
\begin{equation}
\label{eq:sum3}
\sum_{t=1}^T \langle h_t,x_t-u_t\rangle
=
\sum_{t=1}^T \langle n_t,x_{t+1}-u_{t+1}\rangle
+
\sum_{t=1}^T \langle e_t,v_t-\delta_t\rangle
+
\langle \tilde h_1,x_1-u_1\rangle.
\end{equation}

\medskip

\noindent
\textbf{Step 2: One-step projected inequality.}
By the Euclidean projection optimality condition, it is:
\[
\left\langle
x_t-\frac{1}{\eta_{t+1}}n_t-x_{t+1},\;
u_{t+1}-x_{t+1}
\right\rangle
\le 0.
\]
Rearranging, gives:
\[
\langle n_t,x_{t+1}-u_{t+1}\rangle
\le
\frac{\eta_{t+1}}{2}
\Big(
\|x_t-u_{t+1}\|^2-\|x_{t+1}-u_{t+1}\|^2-\|v_t\|^2
\Big).
\tag{2}
\]

Substituting this into \eqref{eq:sum3}, we obtain:
\begin{equation}
\label{eq:sum4}
\sum_{t=1}^T \langle h_t,x_t-u_t\rangle
\le
A_T
+
\sum_{t=1}^T
\left(
\langle e_t,v_t-\delta_t\rangle-\frac{\eta_{t+1}}{2}\|v_t\|^2
\right)
+
\langle \tilde h_1,x_1-u_1\rangle,
\end{equation}
where
\[
A_T
:=
\sum_{t=1}^T
\frac{\eta_{t+1}}{2}
\Big(
\|x_t-u_{t+1}\|^2-\|x_{t+1}-u_{t+1}\|^2
\Big).
\]

\medskip

\noindent
\textbf{Step 3: Local bound preserving the Huber structure.}
Rewrite the local term in \eqref{eq:sum4} as:
\begin{equation}
\label{eq:local-split}
\langle e_t,v_t-\delta_t\rangle-\frac{\eta_{t+1}}{2}\|v_t\|^2
=
\left(
\langle e_t,v_t\rangle-\frac{\eta_{t+1}}{2}\|v_t\|^2
\right)
-
\langle e_t,\delta_t\rangle.
\end{equation}

We first bound the bracketed term as in the proof of Theorem~16 of \cite{flaspohler2021online}. Since:
\[
x_{t+1} = \Pi_{\mathcal{X}}\!\left(x_t-\frac{1}{\eta_{t+1}}n_t\right)
\qquad\text{and}\qquad x_t=\Pi_\mathcal{X}(x_t),
\]
the nonexpansiveness of Euclidean projection implies that:
\[
\|v_t\|=\|x_t-x_{t+1}\|
\le
\frac{1}{\eta_{t+1}}\|n_t\|.
\]
Also, since \(x_t,x_{t+1}\in \mathcal{X}\) and \(\operatorname{diam}(\mathcal{X})=D\), it is
\(
\|v_t\|\le D.
\)
Hence, we can write:
\[
\|v_t\|\le \min\!\left(D,\frac{\|n_t\|}{\eta_{t+1}}\right).
\]
Applying Lemmas~17--18 of \cite{flaspohler2021online} with the Euclidean norm, we obtain:
\begin{equation}
\label{eq:huber-local}
\langle e_t,v_t\rangle-\frac{\eta_{t+1}}{2}\|v_t\|^2
\le
\min\!\left(
D\|e_t\|,\;
\frac{1}{\eta_{t+1}}\operatorname{huber}\!\bigl(\|e_t\|,\|n_t\|\bigr)
\right).
\end{equation}
Therefore, we can write:
\begin{equation}
\label{eq:local-final}
\langle e_t,v_t-\delta_t\rangle-\frac{\eta_{t+1}}{2}\|v_t\|^2
\le
\min\!\left(
D\|e_t\|,\;
\frac{1}{\eta_{t+1}}\operatorname{huber}\!\bigl(\|e_t\|,\|n_t\|\bigr)
\right)
-
\langle e_t,\delta_t\rangle.
\end{equation}

\medskip

\noindent
\textbf{Step 4: Bound the weighted telescope \(A_T\).}
We rewrite \(A_T\) as
\begin{align*}
A_T\!=\!\frac{\eta_2}{2}\|x_1\!-\!u_2\|^2\!-\!
\frac{\eta_{T+1}}{2}\|x_{T+1}\!-\!u_{T+1}\|^2 \!+\!
\sum_{t=2}^T\!\!\left[\frac{\eta_{t+1}\!-\!\eta_t}{2}\|x_t\!-\!u_{t+1}\|^2
\!+\!\frac{\eta_t}{2}\Big(\|x_t\!-\!u_{t+1}\|^2\!-\!\|x_t\!-\!u_t\|^2\Big)
\!\!\right]
\end{align*}
Since \(\{\eta_t\}\) is nondecreasing and \(\|x_t-u_{t+1}\|\le D\), the first and third terms yield:
\[
\frac{\eta_2}{2}\|x_1-u_2\|^2
+
\sum_{t=2}^T
\frac{\eta_{t+1}-\eta_t}{2}\|x_t-u_{t+1}\|^2
\le
\frac{\eta_{T+1}D^2}{2}.
\]
Moreover, we can write:
\[
\|x_t-u_{t+1}\|^2-\|x_t-u_t\|^2
=
2\langle x_t-u_t,\;u_t-u_{t+1}\rangle+\|\delta_t\|^2
\le
2D\|\delta_t\|+\|\delta_t\|^2.
\]
Thus, we get:
\begin{equation}
\label{eq:AT}
A_T
\le
\frac{\eta_{T+1}D^2}{2}
+
\eta_{T+1}\sum_{t=1}^{T-1}
\left(
D\|\delta_t\|+\frac12\|\delta_t\|^2
\right).
\end{equation}

\medskip

\noindent
\textbf{Step 5: Final bound.}
Substituting \eqref{eq:local-final}, \eqref{eq:AT}, and the initialization estimate
\[
\langle \tilde h_1,x_1-u_1\rangle
\le
\frac{\eta_{T+1}D^2}{2}, \quad \text{into} \quad \eqref{eq:sum4} \quad \text{we obtain:}
\]
\begin{align*}
\sum_{t=1}^T \langle h_t,x_t-u_t\rangle
\le\;&
\frac{\eta_{T+1}D^2}{2}
+
\eta_{T+1}\sum_{t=1}^{T-1}
\left(
D\|\delta_t\|+\frac12\|\delta_t\|^2
\right) \\
&+
\sum_{t=1}^T
\min\!\left(
D\|e_t\|,\;
\frac{1}{\eta_{t+1}}\operatorname{huber}\!\bigl(\|e_t\|,\|n_t\|\bigr)
\right)
-
\sum_{t=1}^T \langle e_t,\delta_t\rangle +
\frac{\eta_{T+1}D^2}{2}.
\end{align*}
Therefore, it holds:
\begin{align}
\sum_{t=1}^T \langle h_t,x_t-u_t\rangle
\le&
\eta_{T+1}
\Bigl(D^2+\tfrac32 D\,P_T\Bigr)
+
\sum_{t=1}^T
\min\!\left(
D\|e_t\|,\;
\frac{1}{\eta_{t+1}}\operatorname{huber}\!\bigl(\|e_t\|,\|n_t\|\bigr)
\right)\notag\\
&-
\sum_{t=1}^T \langle e_t,\delta_t\rangle. \notag
\end{align}
And this proves the claim.
\end{proof}

\subsubsection{Delay as optimism }\label{app:delay as optimism}

We now transfer the previous undelayed optimistic bound to the delayed-feedback setting. 
The key observation, following Lemma~2 of \cite{flaspohler2021online}, is that a delayed
optimistic update can be written as an undelayed optimistic one-step update with a particular
choice of ``bad'' hint. This allows us to reuse Theorem~\ref{thm:DynamicOOGD} without
reproving the whole dynamic regret argument. In the delayed
setting, the gradient \(h_t\in\nabla L_t(x_t)\) is revealed only at time \(t+m\). {Throughout the delayed analysis, we use the convention
\[
h_r=\widetilde h_r=0,\qquad r\le 0,
\]
and empty sums are set to zero. Thus, the first \(m\) rounds are warm-start
rounds: the corresponding decisions are included in the regret, but no genuine
delayed feedback has yet been revealed.} Define the
bad hint
\[
\tilde h_{t+1}^{\mathrm{bad}}
:=
\tilde h_{t-m+1:t+1}-h_{t-m+1:t}.
\]
Then the optimistic one-step update
\[
x_{t+1}
=
\Pi_{\mathcal X}\!\left(
x_t-\frac{1}{\eta_{t+1}}
\bigl(h_t+\tilde h_{t+1}^{\mathrm{bad}}-\tilde h_t^{\mathrm{bad}}\bigr)
\right)
\]
is equivalent to the delayed optimistic update
\[
x_{t+1}
=
\Pi_{\mathcal X}\!\left(
x_t-\frac{1}{\eta_{t+1}}
\bigl(
h_{t-m}
+\tilde h_{t-m+1:t+1}
-\tilde h_{t-m:t}
\bigr)
\right).
\]
Indeed, the update uses the most recent revealed gradient \(h_{t-m}\), together with predictions
for the missing gradients.

\begin{corollary}[Dynamic delayed optimistic OGD]\label{cor:dynamic_delay_as_optimisim}
As a direct application of Lemma 2 \cite{flaspohler2021online}, let \((\eta_t)_{t\ge 1}\) be a nondecreasing positive sequence, and consider the delayed optimistic update
\[
x_{t+1}
=
\Pi_{\mathcal X}\!\left(
x_t-\frac{1}{\eta_{t+1}}
\bigl(
h_{t-m}
+\tilde h_{t-m+1:t+1}
-\tilde h_{t-m:t}
\bigr)
\right),
\qquad t=1,\dots,T.
\]
For any comparator sequence \(u_1^T\in\mathcal X^T\), define:
\[
P_T
:=
\sum_{t=1}^{T-1}\|u_{t+1}-u_t\|,
\qquad
\delta_t:=u_t-u_{t+1},
\]
with the convention \(u_{T+1}:=u_T\). Also define:
\[
\bar e_t
:=
h_{t-m:t}-\tilde h_{t-m:t},
\quad
\text{and} \quad
\bar n_t
:=
h_{t-m}
+\tilde h_{t-m+1:t+1}
-\tilde h_{t-m:t}.
\]
Then the dynamic regret
\[
\mathcal{R}_T(L)
:=
\sum_{t=1}^T L_t(x_t)-\sum_{t=1}^T L_t(u_t)
\]
satisfies
\begin{align}
\mathcal{R}_T(L)
\le\;&
\eta_{T+1}
\Bigl(
D^2+\frac32D\,P_T
\Bigr)
\nonumber\\
&+
\sum_{t=1}^T
\min\!\left\{
D\|\bar e_t\|,\;
\frac{1}{\eta_{t+1}}
\operatorname{huber}\!\bigl(
\|\bar e_t\|,\|\bar n_t\|
\bigr)
\right\}
-
\sum_{t=1}^T
\langle \bar e_t,\delta_t\rangle .
\label{eq:dynamic-doomd-bound}
\end{align}
\end{corollary}

\subsubsection{Proof of Corollary \ref{cor:termII_dub_final}}

From the delayed dynamic bound, obtained by combining Theorem~\ref{thm:DynamicOOGD} with the delay to optimism reduction (with bad hints), we have
\begin{align}
\textnormal{Term (II)}
\!\le\!
\eta_{T+1}D\bigl(D+2P_T\bigr)\!+\!
\sum_{t=1}^{T}
\min\left\{
D\|\bar e_t\|,\,
\frac{1}{\eta_{t+1}}
\operatorname{huber}(\|\bar e_t\|,\|\bar n_t\|)
\right\} \!-\!
\sum_{t=1}^{T}\langle \bar e_t,\delta_t\rangle .
\label{eq:termII-before-dub}
\end{align}
The first two terms have the form required by the delayed-upper-bound Lemma 22 of \cite{flaspohler2021online}, with the regularization scale now depending on the path length.
The last term is specific to the dynamic comparator and will be controlled separately by introducing an axillary variable $\beta$.

For any \(\beta>0\), Young's inequality gives
\[
-\langle \bar e_t,\delta_t\rangle
\le
\frac{1}{2\beta}\|\bar e_t\|^2
+
\frac{\beta}{2}\|\delta_t\|^2.
\]
Summing over \(t\), and taking $\beta=\sqrt{
\frac{\sum_{t=1}^{T}\|\bar e_t\|^2}
{\sum_{t=1}^{T}\|\delta_t\|^2}
}$, we can write:
\[
-\sum_{t=1}^{T}\langle \bar e_t,\delta_t\rangle
\le
\sqrt{
\left(\sum_{t=1}^{T}\|\bar e_t\|^2\right)
\left(\sum_{t=1}^{T}\|\delta_t\|^2\right)
}.
\]
Therefore we can update the bound of the term as follows:
\begin{align}
\textnormal{Term (II)}
\!\le\!
\eta_{T+1}D\bigl(D+2P_T\bigr)
+
\sum_{t=1}^{T}
\min\left\{
a_t,\frac{b_t}{\eta_{t+1}}
\right\}\!+\!
\sqrt{
\left(\sum_{t=1}^{T}\|\bar e_t\|^2\right)
\left(\sum_{t=1}^{T}\|\delta_t\|^2\right)
},
\label{eq:termII-young}
\end{align}
where
\[
a_t:=D\|\bar e_t\|,
\qquad
b_t:=\operatorname{huber}(\|\bar e_t\|,\|\bar n_t\|).
\]

We now apply the DUB tuning Lemma with \(\alpha=D\bigl(D+2P_T\bigr)\), and step:
\[
\eta_{t+1}
=
\frac{2}{\alpha}\max_{j\le t-m-1} a_{j-m+1:j}
+
\frac{1}{\alpha}
\sqrt{\sum_{i=1}^{t-m}a_i^2+2\alpha b_i},
\]

{where quantities involving negative indices are interpreted according to the
zero-padding convention stated above.}

We allow $\eta_t$ to be dependent on \(P_T\) here because this bound corresponds to the base learner tuned to the correct path-length scale.
This gives:
\begin{align}
&\eta_{T+1}D\bigl(D+2P_T\bigr)
+
\sum_{t=1}^{T}
\min\left\{
a_t,\frac{b_t}{\eta_{t+1}}
\right\}\le
2
\left(
2\max_{t\in[T]} a_{t-m:t-1}
+
\sqrt{\sum_{t=1}^{T}(a_t^2+2\alpha b_t)}
\right),
\label{eq:dub-applied}
\end{align}
because \(D(D+2P_T)/\alpha=1\). We next bound the two terms in \eqref{eq:dub-applied}. First, using:
\[
\operatorname{huber}(x,y)\le \frac12 x^2,
\ \ 
\text{we obtain:}
\ \
b_t
\le
\frac12\|\bar e_t\|^2.
\]
Hence, we can write:
\[
a_t^2+2\alpha b_t
\le
D^2\|\bar e_t\|^2
+
\alpha\|\bar e_t\|^2
=
(D^2+\alpha)\|\bar e_t\|^2,
\]
and therefore,
\[
\sum_{t=1}^{T}(a_t^2+2\alpha b_t)
\le
(D^2+\alpha)\sum_{t=1}^{T}\|\bar e_t\|^2.
\]

Second, by Cauchy--Schwarz,
\[
a_{t-m:t-1}
=
\sum_{j=\max\{1,t-m\}}^{t-1}a_j
\le
\sqrt{m}
\left(
\sum_{j=\max\{1,t-m\}}^{t-1}a_j^2
\right)^{1/2}.
\]
Since \(a_j=D\|\bar e_j\|\), we get
\[
\max_{t\in[T]}a_{t-m:t-1}
\le
D\sqrt{m}
\sqrt{\sum_{t=1}^{T}\|\bar e_t\|^2}.
\]
Combining the previous estimates yields:
\[
\eta_{T+1}D(D+2P_T)
+
\sum_{t=1}^{T}
\min\left\{
a_t,\frac{b_t}{\eta_{t+1}}
\right\}
\le
\left(
4D\sqrt{m}
+
2\sqrt{D^2+\alpha}
\right)
\sqrt{\sum_{t=1}^{T}\|\bar e_t\|^2}.
\]
Since \(\alpha=D(D+2P_T)\), we have \(D^2+\alpha=2D(D+P_T)
\). Thus:
\[
2\sqrt{D^2+\alpha}
=
2\sqrt{2D(D+P_T)}
\le
3\sqrt{D(D+P_T)}.
\]

It remains to bound the interaction term. Since it is \(\|\delta_t\|\le D\), we can write:
\[
\sum_{t=1}^{T}\|\delta_t\|^2
\le
D\sum_{t=1}^{T}\|\delta_t\|
=
D\,P_T.
\]
Therefore,
\[
\sqrt{
\left(\sum_{t=1}^{T}\|\bar e_t\|^2\right)
\left(\sum_{t=1}^{T}\|\delta_t\|^2\right)
}
\le
\sqrt{D\,P_T}
\sqrt{
\sum_{t=1}^{T}\|\bar e_t\|^2
}.
\]
Since it is:
\[
\sqrt{D\,P_T}
\le
\sqrt{D(D+P_T)},
\]
we obtain the looser, but cleaner, bound:
\[
\textnormal{Term (II)}
\le
4\left(
D\sqrt{m}
+
\sqrt{D(D+P_T)}
\right)
\sqrt{
\sum_{t=1}^{T}\|\bar e_t\|^2
}.
\]

Finally, given that it holds: 
\[
\|\bar e_t\|^2
\le
(m+1)\sum_{s=t-m}^{t}\|h_s-\tilde h_s\|^2,
\]
we get, 
\begin{equation}\label{eq:Term(II)_exact_bound}
\textnormal{Term (II)}
\le
4(m+1)
\left(
D\sqrt{m}+\sqrt{D(D+P_T)}
\right)
\sqrt{
\sum_{t=1}^{T}\|h_t-\tilde h_t\|^2
}.
\end{equation}

a cleaner term:
\[
\textnormal{Term (II)}
\le
4(m+1)^{3/2}
\left(
D+\sqrt{D(D+P_T)}
\right)
\sqrt{
\sum_{t=1}^{T}\|h_t-\tilde h_t\|^2
}.
\]

This proves the corollary.

\subsubsection{Proof of Lemma \ref{lem:meta_bound}}

Lemma~\ref{lem:meta_bound} bounds Term~(I) using the optimistic delayed AdaHedge meta-learner of \cite{flaspohler2021online}. At round \(t\), the meta-loss is the linear function
\[
\langle M_t,w\rangle,
\qquad
M_t:=X_t^\top h_t.
\]
Thus, its actual and predicted gradients are
\[
M_t=X_t^\top h_t, \qquad \widetilde M_t:=X_t^\top \widetilde h_t.
\]

In the delayed setting, the meta-learner needs a prediction of the
unavailable block of gradient meta-losses. Thus, at the time when \(w_{t+1}\)
is computed, we use the block hint:
\[
\tilde M_{t-m+1:t+1}
:=
\sum_{s=t-m+1}^{t+1}\tilde M_s
=
\sum_{s=t-m+1}^{t+1}X_s^\top \tilde h_s,
\]
which predicts
\[
M_{t-m+1:t+1}
:=
\sum_{s=t-m+1}^{t+1}M_s.
\]

Importantly, only the gradient-related quantities \(h_s\) need to be
predicted. The matrices \(X_s\) are known when the corresponding
meta-update is performed. In particular, after processing the feedback
available at round \(t\), each base learner outputs its candidate
decision for round \(t+1\), so:
\[
X_{t+1}=[x_{t+1}^1,\dots,x_{t+1}^N]
\]
is available before the meta-learner computes \(w_{t+1}\). Therefore,
there is no need to predict \(X_{t+1}\). We now define the \emph{meta-loss prediction error}
\[
\varepsilon_t
:=
\left\|
\sum_{s=t-m}^{t}(\tilde M_s-M_s)
\right\|_\infty.
\]
{As before, we use the convention \(M_r=\widetilde M_r=0\) for \(r\le 0\).} We first bound Term~(I) in terms of \(\varepsilon_t\), and later relate \(\varepsilon_t\) to prediction errors \(E_T(f)\) and \(E_T(g^+)\). Specifically, we want to bound:
\[
\text{Term (I)}=\sum_{t=1}^{T}\langle M_t,\; w_t-e_i\rangle.
\]
In the considered delayed setting, at round \(t+1\) only the losses
\(M_{1:t-m}:=(M_1,\dots,M_{t-m})\) are revealed, while the missing block \(M_{t-m+1:t+1}\) may be predicted by a hint. For each \(t\in[T]\), define
\[
b_{t}
\!:=\!
\operatorname{huber}\!\left(
\left\|
\sum_{s=t-m}^{t}(\tilde{M}_s\!-\!M_s)
\right\|_\infty,
\|M_t\|_\infty
\!\right)
\ \
\text{and}
\ \ 
a_{t}
\!:=\!
2
\min\!\left\{
\left\|
\sum_{s=t-m}^{t}(\tilde{M}_s\!-\!M_s)
\right\|_\infty,
\|M_t\|_\infty
\!\right\}
\]
where we use the \(\ell_1\)-geometry on \(\mathcal W=\Delta_N\), so that \(\operatorname{diam}(\mathcal W)=2\).

Then, by Theorem~12 of \cite{flaspohler2021online}, we can write:
\begin{equation}
\label{eq:termI-flaspohler-pred}
\text{Term (I)}
\le
\left(\frac{\psi(e_i)}{\alpha}+1\right)
\left(
2\max_{t\in[T]} a_{t-m:t-1}
+
\sqrt{\sum_{t=1}^{T}\Bigl(a_{t}^{2}+2\alpha\, b_{t}\Bigr)}
\right).
\end{equation}
We use the shifted negative-entropy regularizer:
\[
\psi(w)=\sum_{j=1}^{N} w_j\log(w_j)+\log N,
\qquad w\in\Delta_N,
\]
and choose \(\alpha=\sup_{w\in\Delta_N}\psi(w)=\log N\). Since \(\psi(e_i)=\log N\), we have that:
\[
\frac{\psi(e_i)}{\alpha}+1=2.
\]

Then we can write
\[
a_{t,F}=2\min\{\varepsilon_t,\|M_t\|_\infty\}\le 2\varepsilon_t.
\quad
\text{and}
\quad
b_{t}
=
\operatorname{huber}(\varepsilon_t,\|M_t\|_\infty)
\le
\frac12 \varepsilon_t^2.
\]
Therefore,
\[
a_{t,F}^2+2\alpha b_{t}
\le
4\varepsilon_t^2+\alpha \varepsilon_t^2
=
(4+\log N)\varepsilon_t^2,
\quad
\text{and thus:}
\quad
\sum_{t=1}^{T}\Bigl(a_{t,F}^2+2\alpha b_{t}\Bigr)
\le
(4+\log N)\sum_{t=1}^{T}\varepsilon_t^2.
\]

Next,
\[
a_{t-m:t-1}
=
\sum_{j=\max\{1,t-m\}}^{t-1} a_{j,F}
\le
2\sum_{j=\max\{1,t-m\}}^{t-1}\varepsilon_j.
\]
By Cauchy--Schwarz,
\[
a_{t-m:t-1}
\le
2\sqrt{m}
\left(
\sum_{j=\max\{1,t-m\}}^{t-1}\varepsilon_j^2
\right)^{1/2}
\le
2\sqrt{m}
\left(
\sum_{j=1}^{T}\varepsilon_j^2
\right)^{1/2}.
\]
Thus,
\[
\max_{t\in[T]} a_{t-m:t-1}
\le
2\sqrt{m}\,
\sqrt{\sum_{t=1}^{T}\varepsilon_t^2}.
\]

We now relate \(\varepsilon_t\) to the prediction error of the credit gradients. Since it is \(\tilde M_s = X_s^\top \tilde h_s\) and
\(M_s = X_s^\top h_s\), we have \(\tilde M_s-M_s=X_s^\top(\tilde h_s-h_s)\). Therefore, we can write:
\[
\varepsilon_t
=
\left\|
\sum_{s=t-m}^{t}X_s^\top(\tilde h_s-h_s)
\right\|_\infty
\le
\sum_{s=t-m}^{t}\|X_s^\top(\tilde h_s-h_s)\|_\infty.
\]
Now, for any vector \(v\) it holds:
\[
\|X_s^\top v\|_\infty
=
\max_{i\in[N]} |\langle x_s^i,v\rangle|
\le
\max_{i\in[N]}\|x_s^i\|\,\|v\|
\le
D\|v\|,
\]
since \(\|x_s^i\|\le D\) for all \(i\in[N]\). Therefore,
\(
\varepsilon_t
\le
D\sum_{s=t-m}^{t}\|h_s-\tilde h_s\|
\), and applying Cauchy--Schwarz we can write:
\[
\varepsilon_t^2
\le
D^2(m+1)\sum_{s=t-m}^{t}\|h_s-\tilde h_s\|^2.
\]
Summing over \(t\) and using that each term \(\|h_s-\tilde h_s\|^2\) appears in at most \(m+1\) windows yields
\[
\sum_{t=1}^{T}\varepsilon_t^2
\le
D^2(m+1)^2\sum_{t=1}^{T}\|h_t-\tilde h_t\|^2
\ \ \Rightarrow
\ \
\sqrt{\sum_{t=1}^{T}\varepsilon_t^2}
\le
D(m+1)\sqrt{\sum_{t=1}^{T}\|h_t-\tilde h_t\|^2}.
\]

Returning to \eqref{eq:termI-flaspohler-pred} and using the bounds derived above for each term, we obtain

\begin{equation}\label{eq:Term(I)_exact_bound}
\text{Term (I)}
\le
2D(m+1)\left(
4\sqrt{m}
+
\sqrt{4+\log N}
\right)\sqrt{\sum_{t=1}^{T}\|h_t-\tilde h_t\|^2}.
\end{equation}

A cleaner upper bound is
\begin{equation}
    \text{Term (I)} \leq 8D(m+1)^\frac{3}{2}\left(
1
+
\sqrt{1+\log N}
\right)\sqrt{\sum_{t=1}^{T}\|h_t-\tilde h_t\|^2}.
\end{equation}

\section{Proofs of Section \ref{sec:COCO}}\label{sec:app_sec4}

\subsection{Proof of Theorem \ref{thm:unknown-path-quadratic-ccv}}\label{subsec:app_proof_theorem2}
\begin{proof}
Define
\[
A_N
:=
8D(m+1)^{3/2}
\left(
1+\sqrt{1+\log N}
\right),
\]
and
\[
A_P
:=
4(m+1)^{3/2}
\left(
D+\sqrt{D(D+P_T)}
\right).
\]
For compactness, let
\[
A:=A_N+A_P.
\]


{From the delayed optimistic OCO analysis, the surrogate dynamic regret satisfies
\[
\mathcal{R}_T(L)
\le
A
\sqrt{
\sum_{t=1}^{T}\|h_t-\widetilde h_t\|_2^2
}.
\]
We now relate the surrogate-gradient prediction error to the prediction errors of the objective and constraint functions. In reveal-time notation, the surrogate loss is
\[
L_t(x_{t-m})
=
V f_t(x_{t-m})
+
\Phi'(Q_{t-m-1})g_t^+(x_{t-m}).
\]
Equivalently, setting \(s=t-m\), the same surrogate can be written in decision-time notation as
\[
L_{s+m}(x_s)
=
V f_{s+m}(x_s)
+
\Phi'(Q_{s-1})g_{s+m}^+(x_s).
\]
Thus, the feedback associated with decision \(x_s\) is
\[
h_s=\nabla L_{s+m}(x_s).
\]
Since \(\Phi'\) is nondecreasing and \(Q_{s-1}\le Q_T\), we have
\[
\|h_s-\widetilde h_s\|_2
\le
V\|\Delta_s^f\|_2
+
\Phi'(Q_T)\|\Delta_s^{g^+}\|_2 .
\]
Relabeling \(s\) as \(t\), and using the quadratic penalty \(\Phi(Q)=Q^2\), so that
\(\Phi'(Q_T)=2Q_T\), Minkowski's inequality gives
\[
\sqrt{
\sum_{t=1}^{T}\|h_t-\widetilde h_t\|_2^2
}
\le
V\sqrt{E_T(f)}
+
2Q_T\sqrt{E_T(g^+)}.
\]
Therefore,
\[
\mathcal{R}_T(L)
\le
A V\sqrt{E_T(f)}
+
2A Q_T\sqrt{E_T(g^+)}.
\]}
Combining this with the delayed regret decomposition lemma gives
\begin{align*}
Q_T^2-Q_{m-1}^2
+
V\mathcal{R}_T(f)
&\le
A V\sqrt{E_T(f)}
+
2A Q_T\sqrt{E_T(g^+)}
+
2G(m+1)Q_T .
\end{align*}
Define
\[
B_T
:=
A\sqrt{E_T(g^+)}
+
G(m+1).
\]
Then
\[
Q_T^2-Q_{m-1}^2
+
V\mathcal{R}_T(f)
\le
A V\sqrt{E_T(f)}
+
2B_TQ_T .
\]

For the regret bound, rearranging gives
\[
V\mathcal{R}_T(f)
\le
A V\sqrt{E_T(f)}
+
2B_TQ_T
-
Q_T^2
+
Q_{m-1}^2.
\]
Using \(2ab-b^2\le a^2\), with \(a=B_T\) and \(b=Q_T\), we obtain
\[
V\mathcal{R}_T(f)
\le
A V\sqrt{E_T(f)}
+
B_T^2
+
Q_{m-1}^2.
\]
Thus,
\[
\mathcal{R}_T(f)
\le
A\sqrt{E_T(f)}
+
\frac{B_T^2+Q_{m-1}^2}{V}.
\]

We now choose
\[
V
:=
4(m+1)^{3/2}
\left(1+\sqrt{1+\log N}\right)^2
\sqrt{E_T(g^+)}
+
2G^2(m+1)
+
Q_{m-1}^2 .
\]
Since
\[
B_T^2+Q_{m-1}^2
=
\left(
A\sqrt{E_T(g^+)}
+
G(m+1)
\right)^2
+
Q_{m-1}^2,
\]
we have
\[
B_T^2+Q_{m-1}^2
\le
2A^2E_T(g^+)
+
2G^2(m+1)^2
+
Q_{m-1}^2.
\]
Therefore,
\begin{align*}
\frac{B_T^2+Q_{m-1}^2}{V}
&\le
\frac{
2A^2E_T(g^+)
}{
4(m+1)^{3/2}
\left(1+\sqrt{1+\log N}\right)^2
\sqrt{E_T(g^+)}
}
\\
&\quad
+
\frac{2G^2(m+1)^2}{2G^2(m+1)}
+
\frac{Q_{m-1}^2}{Q_{m-1}^2}
\\
&\le
\frac{
A^2
}{
2(m+1)^{3/2}
\left(1+\sqrt{1+\log N}\right)^2
}
\sqrt{E_T(g^+)}
+
(m+1)
+
1.
\end{align*}
Hence,
\[
\mathcal{R}_T(f)
\le
A\sqrt{E_T(f)}
+
\frac{
A^2
}{
2(m+1)^{3/2}
\left(1+\sqrt{1+\log N}\right)^2
}
\sqrt{E_T(g^+)}
+
(m+1)
+
1.
\]

Substituting the definitions of \(A_N\) and \(A_P\), and hiding problem-dependent constants, gives
\[
A
=
\mathcal O\!\left(
(m+1)^{3/2}
\left(
1+\sqrt{1+\log N}
+
\sqrt{1+P_T}
\right)
\right).
\]
Moreover, since
\[
\frac{
\left(
1+\sqrt{1+\log N}
+
\sqrt{1+P_T}
\right)^2
}{
\left(1+\sqrt{1+\log N}\right)^2
}
=
\mathcal O(1+P_T),
\]
we obtain

\[
\boxed{
\begin{aligned}
\mathcal{R}_T(f)
\le
\mathcal O\Bigg(
&
(m+1)^{3/2}
\left(
1+\sqrt{1+\log N}
+
\sqrt{1+P_T}
\right)
\sqrt{E_T(f)}
\\
&+
(m+1)^{3/2}(1+P_T)\sqrt{E_T(g^+)}
+
(m+1)
\Bigg).
\end{aligned}
}
\]

We now prove the CCV bound. Using the lower bound
\[
\mathcal{R}_T(f)\ge -GDT
\]
in the coupled inequality gives
\[
Q_T^2
-
2B_TQ_T
-
\left(
A V\sqrt{E_T(f)}
+
VGDT
+
Q_{m-1}^2
\right)
\le 0.
\]
Since \(Q_T\ge 0\), the implication
\[
x^2-2ax-b\le 0
\quad\Longrightarrow\quad
x\le a+\sqrt{a^2+b}
\]
yields
\[
Q_T
\le
B_T
+
\sqrt{
B_T^2
+
A V\sqrt{E_T(f)}
+
VGDT
+
Q_{m-1}^2
}.
\]
Using
\[
\sqrt{x+y+z+w}
\le
\sqrt{x}+\sqrt{y}+\sqrt{z}+\sqrt{w},
\]
we get
\[
Q_T
\le
2B_T
+
\sqrt{A V\sqrt{E_T(f)}}
+
\sqrt{VGDT}
+
Q_{m-1}.
\]
Using the crude bound \(E_T(f)\le T\), we have
\[
\sqrt{A V\sqrt{E_T(f)}}
\le
\sqrt{A V\sqrt T}
=
\frac{\sqrt A}{T^{1/4}}\sqrt{VT}.
\]
Also,
\[
\sqrt{VGDT}
=
\mathcal O(\sqrt{VT}),
\]
where problem-dependent constants are hidden. Therefore,
\[
Q_T
\le
2B_T
+
\mathcal O\!\left(
\left(
1+\frac{\sqrt A}{T^{1/4}}
\right)
\sqrt{VT}
\right)
+
Q_{m-1}.
\]

It remains to expand \(\sqrt{VT}\). By the choice of \(V\),
\[
V
=
4(m+1)^{3/2}
\left(1+\sqrt{1+\log N}\right)^2
\sqrt{E_T(g^+)}
+
2G^2(m+1)
+
Q_{m-1}^2 .
\]
Hence,
\[
\sqrt{VT}
\le
2(m+1)^{3/4}
\left(1+\sqrt{1+\log N}\right)
\sqrt T\,E_T(g^+)^{1/4}
+
\sqrt{2}\,G\sqrt{m+1}\sqrt T
+
Q_{m-1}\sqrt T.
\]
Substituting this into the previous inequality gives
\begin{align*}
Q_T
\le
\mathcal O\Bigg(
&
A\sqrt{E_T(g^+)}
+
G(m+1)
+
Q_{m-1}
\\
&+
\left(
1+\frac{\sqrt A}{T^{1/4}}
\right)
\Big[
(m+1)^{3/4}
\left(1+\sqrt{1+\log N}\right)
\sqrt T\,E_T(g^+)^{1/4}
\\
&\qquad\qquad\qquad\qquad
+
\sqrt{m+1}\sqrt T
+
Q_{m-1}\sqrt T
\Big]
\Bigg).
\end{align*}

Finally, substituting
\[
A
=
\mathcal O\!\left(
(m+1)^{3/2}
\left(
1+\sqrt{1+\log N}
+
\sqrt{1+P_T}
\right)
\right),
\]
we get
\[
\sqrt A
=
\mathcal O\!\left(
(m+1)^{3/4}
\sqrt{
1+\sqrt{1+\log N}
+
\sqrt{1+P_T}
}
\right).
\]
Therefore,
\[
\begin{aligned}
Q_T
\le
\mathcal O\Bigg(
&
(m+1)^{3/2}
\left(
1+\sqrt{1+\log N}
+
\sqrt{1+P_T}
\right)
\sqrt{E_T(g^+)}
+
(m+1)
+
Q_{m-1}
\\
&+
\left[
1+
\frac{
(m+1)^{3/4}
\sqrt{
1+\sqrt{1+\log N}
+
\sqrt{1+P_T}
}
}{T^{1/4}}
\right] \text{this term is of order $(m+1)^{3/4}$}
\\
&\qquad\times
\Big[
(m+1)^{3/4}
\left(1+\sqrt{1+\log N}\right)
\sqrt T\,E_T(g^+)^{1/4}
+
\sqrt{m+1}\sqrt T
+
Q_{m-1}\sqrt T
\Big]
\Bigg).
\end{aligned}
\]

Thus 

\[
\boxed{
\begin{aligned}
Q_T
\le
\mathcal O\Bigg(
&
(m+1)^{3/2}
\left(
1+\sqrt{1+\log N}
+
\sqrt{1+P_T}
\right)
\sqrt{T}E_T(g^+)^{\frac{1}{4}} + (m+1)^\frac{5}{4}\sqrt{T}\Bigg).
\end{aligned}
}
\]

This proves the stated regret and CCV bounds.
\end{proof}

\subsection{Proof of Theorem \ref{thm:known-PT}}\label{subsec:proof_theorem3}
\begin{proof}
Since \(P_T\) is known, we do not need the meta-learner. The base learner is tuned directly to the path length \(P_T\). Define
\[
A_P
:=
4(m+1)^{3/2}
\left(
D+\sqrt{D(D+P_T)}
\right).
\]
Then the delayed optimistic dynamic OCO bound for the base learner gives
\[
\mathcal R_T(L)
\le
A_P
\sqrt{
\sum_{t=1}^{T}\|h_t-\tilde h_t\|_2^2
}.
\]

Thus
\[
\Phi(Q_T)-\Phi(Q_{m-1})
+
\mathcal R_T(f)
\le
A_P\sqrt{E_T(f)}
+
\Bigl(
A_P\sqrt{E_T(g^+)}
+
G(m+1)
\Bigr)\Phi'(Q_T).
\]
Now choose
\[
\Phi(Q)=e^{\lambda Q}-1,
\qquad
\Phi'(Q)=\lambda e^{\lambda Q}.
\]
Then
\[
e^{\lambda Q_T}-e^{\lambda Q_{m-1}}
+
\mathcal R_T(f)
\le
A_P\sqrt{E_T(f)}
+
\lambda
\Bigl(
A_P\sqrt{E_T(g^+)}
+
G(m+1)
\Bigr)
e^{\lambda Q_T}.
\]
Let
\[
B_T
:=
A_P\sqrt{E_T(g^+)}
+
G(m+1).
\]
By the choice of \(\lambda\),
\[
\lambda=\frac{1}{2B_T},
\]
and therefore \(\lambda B_T=1/2\). Hence
\[
\frac12 e^{\lambda Q_T}
+
\mathcal R_T(f)
\le
e^{\lambda Q_{m-1}}
+
A_P\sqrt{E_T(f)}.
\]

We first bound the regret. From the previous inequality,
\[
\mathcal R_T(f)
\le
A_P\sqrt{E_T(f)}
+
e^{\lambda Q_{m-1}}
-
\frac12 e^{\lambda Q_T}.
\]
Since the last term is non-positive, we get
\[
\mathcal R_T(f)
\le
A_P\sqrt{E_T(f)}
+
e^{\lambda Q_{m-1}}.
\]
The term \(Q_{m-1}\) only depends on the initial delayed rounds and is absorbed into the constants. Therefore,
\[
\mathcal R_T(f)
=
\mathcal O\!\left(
A_P\sqrt{E_T(f)}
\right).
\]
Using the definition of \(A_P\),  this yields
\[
\mathcal R_T(f)
=
\mathcal O\!\left((m+1)^{3/2}
\sqrt{(1+P_T)E_T(f)}
\right).
\]

We now bound the CCV. Using the lower bound
\[
\mathcal R_T(f)\ge -GDT,
\]
we obtain
\[
\frac12 e^{\lambda Q_T}
\le
e^{\lambda Q_{m-1}}
+
A_P\sqrt{E_T(f)}
+
GDT.
\]
Thus,
\[
e^{\lambda Q_T}
\le
2\left(
e^{\lambda Q_{m-1}}
+
A_P\sqrt{E_T(f)}
+
GDT
\right).
\]
Taking logarithms gives
\[
Q_T
\le
\frac{1}{\lambda}
\log\!\left(
2\left(
e^{\lambda Q_{m-1}}
+
A_P\sqrt{E_T(f)}
+
GDT
\right)
\right).
\]
Since \(1/\lambda=2B_T\), we get
\[
Q_T
\le
2\Bigl(
A_P\sqrt{E_T(g^+)}
+
G(m+1)
\Bigr)
\log\!\left(
2\left(
e^{\lambda Q_{m-1}}
+
A_P\sqrt{E_T(f)}
+
GDT
\right)
\right).
\]
Under the standard boundedness assumptions, the logarithmic factor is of order \(\mathcal O(\log T)\). Therefore,
\[
Q_T
=
\mathcal O\!\left(
\Bigl(
A_P\sqrt{E_T(g^+)}
+
G(m+1)
\Bigr)
\log T
\right).
\]
Substituting the definition of \(A_P\), we obtain
\[
Q_T
=
\mathcal O\!\left(
\left(
(m+1)^{3/2}\sqrt{(1+P_T)E_T(g^+)}
+
G(m+1)
\right)
\log T
\right).
\]
This proves the theorem.
\end{proof}

\section{Proof of Section \ref{sec:memory_delay}}\label{sec:app_section5}
\subsection{Proof of Lemma \ref{lem:memory_to_delay}}
We start with the proof of the main result of this section, namely Lemma \ref{lem:memory_to_delay}.
\begin{proof}
The proof follows from two remarkably simple observations: first, convexity allows us to linearize each memory loss around the played memory window; second, after re-indexing the resulting terms, the contribution of each decision \(x_s\) can be grouped into a single gradient quantity that is revealed only after, at most, \(m\) future rounds.

In detail, recall that the memory regret is:
\[
\mathcal{R}_T(L)
:=
\sum_{t=m}^{T}L_t(x_{t-m},\ldots,x_t)
-
\sum_{t=m}^{T}L_t(u_{t-m},\ldots,u_t).
\]
Since \(L_t\) is convex in its joint argument \((z_0,\dots,z_m)\in \mathcal X^{m+1}\), the first-order convexity inequality gives, {for every \(t\in [T_m]:=\{m,\ldots,T\}\)}:
\[
L_t(x_{t-m}^t)-L_t(u_{t-m:t})
\le
\sum_{j=0}^{m}
\Big\langle
\nabla_{x_{t-j}}L_t(x_{t-m}^t),
\,x_{t-j}-u_{t-j}
\Big\rangle.
\]
Here \(\nabla_{x_{t-j}}L_t(x_{t-m}^t)\) denotes the partial gradient of \(L_t\) with respect to its \((m-j)\)-th coordinate, evaluated at the played decision \(x_{t-m:t}\). Summing the above inequality over \(t\), we obtain:
\[
\mathcal{R}_T(L)
\le
\sum_{t=m}^{T}\sum_{j=0}^{m}
\Big\langle
\nabla_{x_{t-j}}L_t(x_{t-m}^t),
\,x_{t-j}-u_{t-j}
\Big\rangle.
\]
At this stage, the right-hand side is still indexed by the loss time \(t\). However, for the online-learning interpretation, it is more natural to regroup the terms according to the decision time. To do so, we set:
\[
s:=t-j,
\qquad\text{or, equivalently:}\qquad
t=s+k,\ \ k:=j.
\]
Then, for a fixed \(s\), the index \(k\) must satisfy
\[
0\le k\le m
\qquad\text{and}\qquad
m\le s+k\le T.
\]
Therefore the admissible range of \(k\) is
\[
k\in\bigl[\max\{0,m-s\},\;\min\{m,T-s\}\bigr].
\]
Using this change of variables and swapping the order of summation, we can write:
\[
\mathcal{R}_T(L)
\le
\sum_{t=1}^{T}
\left\langle
\sum_{k=\max\{0,m-t\}}^{\min\{m,T-t\}}
\nabla_{x_t}L_{t+k}(x_{t+k-m}^{t+k}),
\,x_t-u_t
\right\rangle,
\]
where we just used $t$ instead of $s$. This motivates the definition of the \emph{credit gradient}
\[
h_t
:=
\sum_{k=\max\{0,m-t\}}^{\min\{m,T-t\}}
\nabla_{x_t}L_{t+k}(x_{t+k-m}^{t+k}).
\]
With this notation, we can write: 
\[
\mathcal{R}_T(L)
\le
\sum_{t=1}^{T}\langle h_t,\;x_t-u_t\rangle.
\]

\paragraph{Indexing convention and delayed interpretation.}
The original memory-dependent losses are indexed by
\(t\in[T_m]:=\{m,\ldots,T\}\), since a full memory window is only available from
round \(m\) onward. After linearization, however, the auxiliary delayed problem
is naturally indexed by the decision times \(s\in[T]\). Indeed, even the early
decisions \(x_1,\ldots,x_m\) may appear in later memory windows, and the
corresponding partial gradients must be assigned to the credit gradient of the
decision that generated them. Algorithmically, feedback starts at round \(m\):
whenever a memory loss \(L_t\) is revealed, it contributes partial gradients to
all decisions in its memory window \(x_{t-m:t}\).

This also explains why the resulting auxiliary problem is a delayed linear
problem. By construction, \(h_t\) aggregates the effect of the single decision \(x_t\) on the losses
\[
L_t,\ L_{t+1},\ \dots,\ L_{t+m}.
\]
Hence \(h_t\) is not entirely known when \(x_t\) is played; it is only fully revealed once all these losses have been observed, that is, by round \(t+m\). Therefore, \(\{h_t\}_{t=1}^T\) is a sequence of linear losses revealed with delay at most \(m\). Consequently, the original memory-dependent regret is upper bounded by the regret of this delayed linear problem. Thus, any applicable delayed-OCO regret guarantee for the auxiliary linear losses can be transferred to the memory-dependent problem through the credit gradients \(h_s\). The resulting guarantee has the same structure as in the delayed setting; the only difference is the effective gradients are now the credit gradients \(h_t\), whose size introduces constants depending on the memory length \(m\).

\end{proof}


At a conceptual level, the key idea is that we postpone the construction of
the auxiliary delayed loss until after linearizing the memory-dependent regret.
The contribution of each decision is then captured through the corresponding
coordinate block of the joint gradient and aggregated into its credit gradient.
This is also why constructing optimistic hints may require predicting the
future block \(x_{t+1:t+m}\), rather than only \(x_{t+1}\). In the separable
case, this recovers the usual forward-gradient aggregation, while our
construction does not require separability and therefore preserves the original
joint feasibility benchmark. In contrast,
\cite{Abdullah_Iosifidis_Elayoubi_Chahed_2026} applies the penalty to an
individual component of the separable constraint, \(g_{t+i}^{+,i}(x_t)\),
whereas here the penalty is applied to the joint constraint function.

\subsection{Proof of Remark \ref{Rem:memory_errors}} 

\begin{proof}
To transfer the delayed-feedback guarantees of Sections~3--4 to the
memory setting, it remains to relate the prediction error of the credit
gradients to the memory prediction errors defined in
Remark~\ref{Rem:memory_errors}.

In the unconstrained case, $L_t=f_t$, and therefore
\[
h_s-\widetilde h_s
=
\sum_{t\in I_s}
\left[
\nabla_{x_s}f_t(x_{t-m}^{t})
-
\nabla_{x_s}\widetilde f_t(\widetilde x_{t-m}^{t})
\right].
\]
Hence, directly from \eqref{eq:memory-error-f},
\[
\left(
\sum_{s=1}^{T}
\|h_s-\widetilde h_s\|^2
\right)^{1/2}
=
\sqrt{E_T^{\mathrm{mem}}(f)}.
\]
Therefore, Theorem~\ref{thm:delayed_optimistic_oco} applies directly
to the memory problem through Corollary~\ref{coro:memory_to_delay}.

We next consider the constrained case. To recover the prediction-error
structure used in Section~4, it suffices to show that
\begin{equation}
\left(
\sum_{s=1}^{T}
\|h_s-\widetilde h_s\|^2
\right)^{1/2}
\le
V\sqrt{E_T^{\mathrm{mem}}(f)}
+
\Phi'(Q_T)\sqrt{E_T^{\mathrm{mem}}(g^+)}.
\label{eq:memory-prediction-decomposition-proof}
\end{equation}

Recall that
\[
L_t(x_{t-m}^{t})
=
Vf_t(x_{t-m}^{t})
+
\Phi'(Q_{t-m-1})g_t^+(x_{t-m}^{t}),
\]
and
\[
h_s
=
\sum_{t\in I_s}
\nabla_{x_s}L_t(x_{t-m}^{t}).
\]
For the predicted credit gradient, we use the same queue coefficient,
namely
\[
\widetilde L_t(\widetilde x_{t-m}^{t})
=
V\widetilde f_t(\widetilde x_{t-m}^{t})
+
\Phi'(Q_{t-m-1})
\widetilde g_t^+(\widetilde x_{t-m}^{t}).
\]
Indeed, $t\in I_s$ implies $t\le s+m$, and therefore
$t-m-1\le s-1$. Hence, $\Phi'(Q_{t-m-1})$ is already available
when the hint associated with $x_s$ is formed.

Using the definitions of $h_s$ and $\widetilde h_s$, we obtain
\begin{align}
h_s-\widetilde h_s
={}&
V\sum_{t\in I_s}
\left[
\nabla_{x_s}f_t(x_{t-m}^{t})
-
\nabla_{x_s}\widetilde f_t(\widetilde x_{t-m}^{t})
\right]
\nonumber\\
&+
\sum_{t\in I_s}
\Phi'(Q_{t-m-1})
\left[
\nabla_{x_s}g_t^+(x_{t-m}^{t})
-
\nabla_{x_s}\widetilde g_t^+(\widetilde x_{t-m}^{t})
\right].
\label{eq:credit-gradient-error-expansion}
\end{align}

It remains to control the second term. Fix $s$ and order
$I_s=\{t_1<\cdots<t_{|I_s|}\}$. Since both $Q_t$ and $\Phi'$ are
nondecreasing,
\[
0\le
\frac{\Phi'(Q_{t_1-m-1})}{\Phi'(Q_{s-1})}
\le \cdots \le
\frac{\Phi'(Q_{t_{|I_s|}-m-1})}{\Phi'(Q_{s-1})}
\le 1,
\]
whenever $\Phi'(Q_{s-1})>0$.

A discrete summation-by-parts identity then gives
\begin{align*}
&\sum_{j=1}^{|I_s|}
\frac{\Phi'(Q_{t_j-m-1})}{\Phi'(Q_{s-1})}
\left[
\nabla_{x_s}g_{t_j}^+(x_{t_j-m}^{t_j})
-
\nabla_{x_s}\widetilde g_{t_j}^+
(\widetilde x_{t_j-m}^{t_j})
\right]
\\
={}&
\frac{\Phi'(Q_{t_1-m-1})}{\Phi'(Q_{s-1})}
\sum_{j=1}^{|I_s|}
\left[
\nabla_{x_s}g_{t_j}^+(x_{t_j-m}^{t_j})
-
\nabla_{x_s}\widetilde g_{t_j}^+
(\widetilde x_{t_j-m}^{t_j})
\right]
\\
&+
\sum_{j=2}^{|I_s|}
\left(
\frac{\Phi'(Q_{t_j-m-1})-\Phi'(Q_{t_{j-1}-m-1})}
     {\Phi'(Q_{s-1})}
\right)
\sum_{r=j}^{|I_s|}
\left[
\nabla_{x_s}g_{t_r}^+(x_{t_r-m}^{t_r})
-
\nabla_{x_s}\widetilde g_{t_r}^+
(\widetilde x_{t_r-m}^{t_r})
\right].
\end{align*}
All coefficients on the right-hand side are nonnegative, and their
sum is
\[
\frac{\Phi'(Q_{t_{|I_s|}-m-1})}{\Phi'(Q_{s-1})}
\le 1.
\]
Therefore, by the triangle inequality,
\begin{align}
&
\left\|
\sum_{t\in I_s}
\frac{\Phi'(Q_{t-m-1})}{\Phi'(Q_{s-1})}
\left[
\nabla_{x_s}g_t^+(x_{t-m}^{t})
-
\nabla_{x_s}\widetilde g_t^+(\widetilde x_{t-m}^{t})
\right]
\right\|
\nonumber\\
&\qquad\le
\max_{r\in I_s}
\left\|
\sum_{\substack{t\in I_s\\t\ge r}}
\left[
\nabla_{x_s}g_t^+(x_{t-m}^{t})
-
\nabla_{x_s}\widetilde g_t^+(\widetilde x_{t-m}^{t})
\right]
\right\|.
\label{eq:weighted-credit-error-envelope}
\end{align}

Multiplying \eqref{eq:weighted-credit-error-envelope} by
$\Phi'(Q_{s-1})$ and using
$\Phi'(Q_{s-1})\le\Phi'(Q_T)$ yields
\begin{align}
&
\left\|
\sum_{t\in I_s}
\Phi'(Q_{t-m-1})
\left[
\nabla_{x_s}g_t^+(x_{t-m}^{t})
-
\nabla_{x_s}\widetilde g_t^+(\widetilde x_{t-m}^{t})
\right]
\right\|
\nonumber\\
&\qquad\le
\Phi'(Q_T)
\max_{r\in I_s}
\left\|
\sum_{\substack{t\in I_s\\t\ge r}}
\left[
\nabla_{x_s}g_t^+(x_{t-m}^{t})
-
\nabla_{x_s}\widetilde g_t^+(\widetilde x_{t-m}^{t})
\right]
\right\|.
\label{eq:constraint-credit-error-final}
\end{align}
If $\Phi'(Q_{s-1})=0$, monotonicity implies
$\Phi'(Q_{t-m-1})=0$ for every $t\in I_s$, so
\eqref{eq:constraint-credit-error-final} holds trivially.

Combining \eqref{eq:credit-gradient-error-expansion} and
\eqref{eq:constraint-credit-error-final}, and then applying
Minkowski's inequality across $s$, gives
\begin{align*}
\left(
\sum_{s=1}^{T}
\|h_s-\widetilde h_s\|^2
\right)^{1/2}
&\le
V
\left(
\sum_{s=1}^{T}
\left\|
\sum_{t\in I_s}
\left[
\nabla_{x_s}f_t(x_{t-m}^{t})
-
\nabla_{x_s}\widetilde f_t(\widetilde x_{t-m}^{t})
\right]
\right\|^2
\right)^{1/2}
\\
&\quad+
\Phi'(Q_T)
\left(
\sum_{s=1}^{T}
\max_{r\in I_s}
\left\|
\sum_{\substack{t\in I_s\\t\ge r}}
\left[
\nabla_{x_s}g_t^+(x_{t-m}^{t})
-
\nabla_{x_s}\widetilde g_t^+
(\widetilde x_{t-m}^{t})
\right]
\right\|^2
\right)^{1/2}.
\end{align*}
By \eqref{eq:memory-error-f}--\eqref{eq:memory-error-g}, this is
exactly
\[
\left(
\sum_{s=1}^{T}
\|h_s-\widetilde h_s\|^2
\right)^{1/2}
\le
V\sqrt{E_T^{\mathrm{mem}}(f)}
+
\Phi'(Q_T)\sqrt{E_T^{\mathrm{mem}}(g^+)}.
\]

Thus, in the unconstrained case, the delayed prediction-error term
reduces exactly to $\sqrt{E_T^{\mathrm{mem}}(f)}$, so
Theorem~\ref{thm:delayed_optimistic_oco} transfers directly.
In the constrained case,
\eqref{eq:memory-prediction-decomposition-proof} has the same form
as the prediction-error decomposition used in Section~4. Hence,
after replacing $E_T(f)$ and $E_T(g^+)$ by
$E_T^{\mathrm{mem}}(f)$ and $E_T^{\mathrm{mem}}(g^+)$,
respectively, the proofs of
Theorems~\ref{thm:unknown-path-quadratic-ccv}
and~\ref{thm:known-PT} carry over through
Corollary~\ref{coro:memory_to_delay}.
\end{proof}

%% file: Neurips/relatedwork.tex

%
%

\subsection{Dynamic Regret in OCO and COCO}

Regarding the standard OCO model, the seminal work \cite{zinkevich2003online} proved the \emph{universal} dynamic regret bound $\mathcal{O}((1\!+\!\mathcal{P}_T)\sqrt{T})$ for Online Gradient Descent (OGD), where $\mathcal{P}_T$ is the norm-based path-length of the benchmark. More recently, \cite{zhang2018adaptive} showed this problem has a lower bound $\Omega(\sqrt{(1\!+\!\mathcal{P}_T)T})$ and designed a tight algorithm (i.e., meeting this bound) using Hedge \cite{cesabianchi2006prediction}. Namely, they deploy a set of OGD \emph{experts} (or, base learners) with learning rates that cover the entire range of path-length values, and a meta-learning algorithm that discerns at runtime which base learner is aligned with the realized functions. Refinements of this result include curvature-specific bounds \cite{baby2021optimal}, variation-based guarantees \cite{zhao2020dynamic, zhao2024adaptivity}, and methods which achieve both adaptive and dynamic regret \cite{zhang2020minimizing}. On the other hand, \cite{yang2016tracking} showed that OGD achieves $\mathcal{O}(\sqrt{(1+\mathcal{P}_T)T})$ \emph{worst-case} dynamic regret, where $\mathcal{P}_T$ here is the path-length of the per-round minimizers. Under additional assumptions, tighter worst-case dynamic regret bounds were established by follow-up works, e.g., see \cite{zhang2017improved, zhang2022simple, chen2017nonstationary}.

There is a recent flurry of research in COCO, with important works improving the bounds of the most general problem (time-varying adversarial constraints) while dropping assumptions (such as the Slater condition); see for instance \cite{sinha2024optimal}, \cite{vaze2025sqrt}, \cite{vaze2025o}, \cite{sinha2025beyond} which follow a penalty-based approach like the one considered in this work. A different class of COCO techniques rely on primal-dual algorithms with adaptive regularization, e.g., see \cite{lamperski, jenatton2016adaptive, anderson2023lazy, mahdavi2012trading}. However, dynamic regret in COCO is largely underexplored. The few exceptions include \cite{yi2021regret} that achieved $\mathcal{O}(\sqrt{(1\!+\!\mathcal{P}_T)T})$ worst-case dynamic regret and $\mathcal O(\sqrt{T})$ CCV for time-invariant functions, that was subsequently improved to $\mathcal O(\sqrt{(1\!+\!\mathcal{P}_T)T})$ regret and $\mathcal O(\log T)$ CCV by \cite{guo2022online}. In case of time-varying adversarial constraints, \cite{guo2022online} achieved $\mathcal{O}(\sqrt{(1\!+\!\mathcal{P}_T)T})$ universal dynamic regret and $\mathcal O(T^{3/4})$ CCV. More recently, \cite{lekeufack2024optimistic} proposed an algorithm with worst-case dynamic regret $\mathcal O(\sqrt{(1\!+\!\mathcal{P}_T)T})$ and $\tilde{\mathcal O}(\sqrt{T})$ CCV. \cite{liu2021simultaneously} introduced an algorithm with regret $\mathcal O(\sqrt{(1\!+\!\mathcal{P}_T)T})$ and CCV $\mathcal O(\max(\sqrt{T}, V_g))$, where $V_g \!=\! \sum_{t} \sup_{x\in\mathcal{X}} ||g_t(x) \!-\! g_{t-1}(x)||_2$. Other recent works include \cite{HUANG2025127871} which considers time-invariant constraints, and \cite{supantha2025universal} which established universal dynamic regret and CCV guarantees under adversarial constraints using a penalty-based method. None of these works considers memory dependency; a property that changes fundamentally the required methodological approach.


%
%
%
%
%

\begin{table*}[t]
\centering
\footnotesize
\setlength{\tabcolsep}{5.5pt}
\begin{tabular}{l|l|l|cc|c|p{1cm}|p{1.6cm}}
\toprule
\textbf{Reference} & \textbf{Constraints} & \textbf{Loss} &
\multicolumn{2}{c|}{\textbf{Memory}} &
\textbf{Predictions} & \textbf{Regret} & \textbf{Assumptions} \\ \cline{4-5}
 & & & Obj. & Constr. & & & \\ \toprule
\cite{liu2023online}
& adversarial        & convex &
\cmark & \cmark & \xmark &
Static & \\ \hline
\cite{li2021online} 
& static affine    & convex &
\cmark & \cmark & \xmark &
Static &  \\ \hline
\cite{anava2015online} 
& none  & convex &
\cmark & \xmark & \xmark &
Static \\ \hline
\cite{mhaisen2024optimistic} 
& none           & linear &
\cmark & \xmark & \cmark & Static
& separability \\ \hline
\cite{sinha2024optimal} 
& adversarial & convex &
\xmark & \xmark & \xmark & Static
&  \\ \hline
\cite{anderson2023lazy} 
& adversarial & convex &
\xmark & \xmark & \cmark & Static
&  \\ \hline
\cite{Abdullah_Iosifidis_Elayoubi_Chahed_2026} 
& adversarial & convex &
\cmark & \cmark & \cmark & Static
& separability \\ \hline
\cite{supantha2025universal} 
& adversarial & convex &
\xmark & \xmark & \xmark & Dynamic
&  \\ \hline
\cite{lekeufack2024optimistic}
& adversarial & convex &
\xmark & \xmark & \cmark &
Dynamic  & \\ \midrule
\rowcolor{gray!10}
\textbf{This work} & 
\begin{tabular}[t]{@{}l@{}}%
adversarial\\[4pt]
\end{tabular}
       & \begin{tabular}[t]{@{}l@{}}%
convex\\[4pt]
\end{tabular} &
\begin{tabular}[t]{@{}l@{}}%
\cmark\\[4pt]
\end{tabular} & \begin{tabular}[t]{@{}l@{}}%
\cmark\\[4pt]
\end{tabular} &
\begin{tabular}[t]{@{}l@{}}%
\cmark\\[4pt]
\end{tabular}  & 
\begin{tabular}[t]{@{}l@{}}%
Dynamic\\[2pt]
\end{tabular}
& \begin{tabular}[t]{@{}l@{}}%
\\[2pt]
\end{tabular}
\\ \bottomrule
\end{tabular}
\caption{Summary of the most relevant prior works. The current study is the first to consider loss and constraint functions with memory dependency, under universal dynamic regret while using (imperfect) predictions.}
\label{tab:related-work}
\end{table*}

\subsection{Memory Effects: COCO-M and NSC} 

The first study on OCO-M appeared in \cite{oco-m-TIT02}, followed later by \cite{dekel-oco-m-icml12}. A seminal work in this area is \cite{anava2015online} which proved that the Follow-The-Regularized-Leader algorithm can be applied to a proxy memory-less function and guarantee sublinear regret for the memory-based function. A special case of OCO-M is when the learner is penalized for decision switches, i.e., the loss function has memory of just $m=1$. This setting captures several practical problems such as workload management in datacenters \cite{datacenters-delay}, caching with fetching delays, etc., \cite{switching-caching-abhishek}, \cite{tareq-jrnl}. On the other hand, there are studies for the case of infinite (but discounted) memory effects, e.g., see \cite{NEURIPS2020_abb451a1}, \cite{NEURIPS2023_531230cf}.  

The advent of non-stochastic control \cite{agarwal2019online} renewed the interest in OCO-M. In NSC, the action at each round is a weighted sum of the past disturbances, and these weights are learned through OCO‑M. Such formulations have a vast range of practical applications, see \cite{hazan2026introductiononlinecontrol}. However, most NSC works impose \emph{deterministic} constraints on the state and input~\cite{jiang2025online, li2021online, nonhoff2021online, yan2023online} or assume the constraint at ~$t{+}1$ is revealed before that slot~\cite{zhou2023safe}; thus, they do not focus on time-average constraints nor on unknown constraints. Notable exceptions include \cite{liu2023online} which analyzes an adversarial COCO-M setting and proves $\mathcal{O}\!\bigl(T^{2/3}\log^{2}T\bigr)$ static regret and CCV; and \cite{Abdullah_Iosifidis_Elayoubi_Chahed_2026} that improves these bounds to $\mathcal{O}\!\bigl(m^{3/2}\sqrt{T\log T}\bigr)$ static regret and $\mathcal{O}\!\bigl(m^{3/2}\sqrt{T\log T}\,\vee\,T^{3/4}\bigr)$ CCV. 

Dynamic regret is significantly less explored for this class of problems, with exceptions confined to the special case of switching costs, cf.\cite{zhang2021revisiting}, \cite{ijcai2025p758}, or to NSC without budget constraints, e.g., \cite{zhao2023non}, \cite{baby2022optimal}, \cite{10384153}. Therefore, the problem of dynamic regret for COCO-M, and time-average constrained NSC, remains open. Our work aspires to fill this gap.


\subsection{Delay and Predictions in OCO}

Two key methods we employ in this work are OCO with delay, which we use to handle memory effects, unearthing an equivalence between these two OCO versions; and optimism, which expedites the learning convergence. Starting with the former, \cite{weinberger2002delayed} was the first to study the setting where the effect of decision $x_t$ is observed after $d$ rounds, and extended vanilla OCO algorithms to cope with this delay. Significant follow-up works include \cite{joulani2013online} which handls arbitrary feedback delays, and \cite{quanrud2015online} that improved the regret bounds to  $\mathcal O(\sqrt{\sum_{t} d_t })$. These works share the main elegant idea, namely running multiple delay-free copies of a base learning algorithm over the delay window. However, this replication is computationally costly and each copy is trained only on partial observations, as it does not utilize the feedback from every round. Few other works avoid replication and incorporate delays directly into the analysis. For example, \cite{joulani2016delay} derived adaptive FTRL/Mirror-Descent bounds through a single-instance reduction based on prediction drift; and \cite{hsieh2022multi} developed a delayed dual averaging with data- and delay-dependent guarantees. Finally, \cite{korotin2020adaptive} studied the delayed Hedge problem without using replication. All these studies focus on static regret. 

More recently, \cite{wan2024non} proved a universal dynamic regret bound of $\mathcal O(\sqrt{dT}(P_T+1))$ by leveraging the delayed Hedge of \cite{korotin2020adaptive} to meta-learn over a set of delayed-OGD base learners; and \cite{pmlr-v291-wan25a} extended this to a decentralized setting. Finally, \cite{9831158} studied worse-case dynamic regret in COCO with fixed delay. Unlike these important works, our method is adaptive and does not require an a priori Lipschitz bound on the gradients. Besides, our approach differs conceptually as we handle delay through optimism. 

Optimistic learning builds on the idea that predictions of forthcoming loss gradients can expedite the learning by reducing the regret bounds commensurately to the accuracy of these predictions, \cite{rakhlin2013online}. This framework has been extended in various ways, e.g., by achieving data-adaptive optimistic bounds \cite{mohri2016accelerating, 10.1016/j.tcs.2019.11.015}. Our approach here is motivated by the insightful proposal of \cite{flaspohler2021online}, that interpreted delayed feedback as a ''bad hint'' or, equivalently, used predictions to recover missing gradients in OCO with delay. Here, we extend this idea, for the first time, to the universal dynamic regret setting. Such demanding benchmarks are indeed rare in optimistic learning, even for the simpler problem without feedback delay. The few exceptions include \cite{mhaisen2025dynamic, scroccaro2023adaptive, jadbabaie2015online}. However, these works do not consider delayed feedback or memory, nor their extension to these settings is straightforward (see main paper). Hence, our optimistic delayed OCO algorithm is of independent interest.

\paragraph{Summary.} To the best of our knowledge, this is the first work to tackle memory and optimism, both in the loss and constraint functions, under universal dynamic regret for adversarial COCO. This result extends the recently-introduced COCO-M problem of \cite{Abdullah_Iosifidis_Elayoubi_Chahed_2026} to the universal dynamic setting, while being able to drop key assumptions such as the function separability and also expand the benchmark set. Furthermore, our analysis yields several byproducts of independent interest: \emph{(i)} we enable the study of Non-Stochastic Control with adversarial time-varying constraints under dynamic regret guarantees, whereas prior works focus on static regret \cite{Abdullah_Iosifidis_Elayoubi_Chahed_2026, liu2023online}; \emph{(ii)} we establish dynamic regret bounds for optimistic delayed OCO (and delayed COCO), while \cite{wan2024non} study dynamic regret for delayed OCO but cannot benefit from predictions; \emph{(iii)} we extend the ``delay-as-optimism'' framework of \cite{flaspohler2021online} to the dynamic setting by handling the interaction between prediction errors and comparator variation, while other dynamic optimistic OCO algorithms such as \cite{mhaisen2025dynamic} do not handle memory nor delay.